%% file: main.tex
\documentclass[twoside]{article}

\usepackage[accepted]{aistats2023}
\usepackage[round]{natbib}

\usepackage{amsfonts,amsthm,amssymb}
\usepackage{mathtools}
\usepackage{graphicx}
\usepackage{subcaption}
\usepackage{hyperref}
\usepackage{xcolor}
\usepackage{verbatim}

\usepackage{xspace}
\usepackage{multirow}
\usepackage{booktabs}

\input{definitions}

\usepackage{color}

\begin{document}

%
\runningtitle{Convex Bounds on the Softmax Function for Robustness Verification}

%
\runningauthor{Dennis Wei, Haoze Wu, Min Wu, Pin-Yu Chen, Clark Barrett, Eitan Farchi}

\twocolumn[

\aistatstitle{Convex Bounds on the Softmax Function with \\Applications to Robustness Verification}

\aistatsauthor{Dennis Wei \And Haoze Wu \And  Min Wu}

\aistatsaddress{IBM Research \And Stanford University \And Stanford University}

\aistatsauthor{Pin-Yu Chen \And Clark Barrett \And Eitan Farchi}

\aistatsaddress{IBM Research \And Stanford University \And IBM Research} ]

\begin{abstract}
  The softmax function is a ubiquitous component at the output of neural networks and increasingly in intermediate layers as well. This paper provides convex lower bounds and concave upper bounds on the softmax function, which are compatible with convex optimization formulations for characterizing neural networks and other ML models. We derive bounds using both a natural exponential-reciprocal decomposition of the softmax as well as an alternative decomposition in terms of the log-sum-exp function. The new bounds are provably and/or numerically tighter than linear bounds obtained in previous work on robustness verification of transformers. As illustrations of the utility of the bounds, we apply them to verification of transformers as well as of the robustness of predictive uncertainty estimates of deep ensembles.
\end{abstract}

\input{intro}

\input{relWork}

\input{prelim}

\input{ER}

\input{LSE}

\input{synth}

\input{appl}

\input{concl}

\subsubsection*{Acknowledgements}
This work was partially supported by IBM as a founding member of Stanford Institute for Human-centered Artificial Intelligence (HAI).  Additional support was provided by a direct grant from HAI and by the Stanford Center for AI Safety.

\bibliographystyle{plainnat}
\bibliography{main}

\appendix
\onecolumn

\input{appendix}

\vfill

\end{document}

%% file: definitions.tex
\DeclareMathOperator{\se}{SE}
\DeclareMathOperator{\lse}{LSE}
\DeclareMathOperator{\sm}{softmax}
\DeclareMathOperator*{\argmax}{arg\,max}
\newcommand{\given}[1][]{\:#1\vert\:}

\newcommand{\ulv}{\underline{v}}
\newcommand{\olv}{\overline{v}}
\newcommand{\ulp}{\underline{p}}
\newcommand{\olp}{\overline{p}}
\newcommand{\ulq}{\underline{q}}
\newcommand{\olq}{\overline{q}}
\newcommand{\olse}{\overline{\se}}

\newcommand{\tl}{\tilde{l}}
\newcommand{\tu}{\tilde{u}}
\newcommand{\tx}{\tilde{x}}

\newcommand{\cI}{\mathcal{I}}
\newcommand{\cA}{\mathcal{A}}
\newcommand{\cU}{\mathcal{U}}

\newcommand{\lin}{\mathrm{lin}}

\newcommand{\ER}{\mathrm{ER}}

\DeclareMathOperator{\R}{\mathcal{R}}

\newtheorem{theorem}{Theorem}

\newcommand{\dennis}[1]{}

\newcommand{\attSmall}{\texttt{A$\_$small}\xspace}
\newcommand{\attMed}{\texttt{A$\_$med}\xspace}
\newcommand{\attBig}{\texttt{A$\_$big}\xspace}

%% file: intro.tex
\section{INTRODUCTION}
\label{sec:intro}

The softmax function is an indispensable component of multiclass classifiers ranging from multinomial logistic regression models to deep neural networks (NNs). It is most often deployed at the output of a classifier to convert $K$ real-valued scores corresponding to $K$ classes into a probability distribution over the classes. More recently, the softmax is playing an increasing role in intermediate layers as well with the popularization of Transformers~\citep{vaswani2017attention}, whose quintessential component, the (self\nobreakdash-)attention mechanism~\citep{luong2015effective,gehring2017convolutional}, utilizes softmax to compute attention scores.

Our main contribution in this paper is to provide convex bounds on the softmax function. More precisely, we derive lower bounds on the outputs of the softmax that are convex functions of the inputs, and upper bounds on the outputs that are concave functions of the inputs (see \eqref{eqn:LU} later). This enables the formulation of convex optimization problems for characterizing ML models with softmax components, particularly in intermediate layers.

We apply our bounds to verification of the robustness of NNs against adversarial input perturbations. 
We consider in particular the quantification of predictive uncertainty for multiclass classifiers, which is typically assessed in terms of accurate estimation of the conditional probability distribution. We are not aware of prior work that directly addresses the robustness of uncertainty estimation metrics, especially for deep ensembles \citep{lakshminarayanan2017simple,rahaman2021uncertainty}, although the works of \citet{bitterwolf2020certifiably,berrada2021make} are related.

Our results are summarized as follows. In Section~\ref{sec:ER}, we first consider an exponential-reciprocal decomposition of the softmax function, used by \citet{Shi2020Robustness,bonaert2021fast} in 
robustness verification of transformers. While \citet{Shi2020Robustness,bonaert2021fast} limited themselves to linear bounds, we instead derive nonlinear convex bounds (which we refer to as ``ER'') and show that these are tighter than the previous linear bounds (``lin''). We then consider in Section~\ref{sec:LSE} an alternative decomposition in terms of the log-sum-exp (LSE) function, a well-known convex function \citep{boyd2004convex}, and obtain corresponding bounds. We prove that the LSE upper bound is always tighter than the ER upper bound, and that the LSE lower bound is tighter than the ER lower bound for the case of $K=2$ inputs. These analytical results are summarized by the following inequalities, where $L(x)$ and $U(x)$ denote lower and upper bounds that are functions of the input $x$:
\begin{multline}\label{eqn:ineqChain}
    L^\lin(x) \leq L^\ER(x) \underbrace{\leq}_{K=2} L^{\lse}(x) \leq \sm(x)\\
    \leq U^{\lse}(x) \leq U^\ER(x) \leq U^\lin(x).
\end{multline}
For $K > 2$ inputs, while there are instances where $L^\ER(x) > L^{\lse}(x)$, our numerical experiment in Section~\ref{sec:synth} suggests that this does not occur often and that in some regimes, $L^{\lse}$ is tighter by factors of $3$ or more in terms of the mean gap with respect to $\sm(x)$. For the upper bounds, we find that $U^\ER(x)$ improves considerably upon $U^\lin(x)$, which can be rather loose, and $U^{\lse}(x)$ consistently improves upon $U^\ER(x)$ by a further factor of $2$.
In Section~\ref{sec:appl}, we describe experiments on robustness verification of transformers and of uncertainty estimation by deep ensembles. The results provide further evidence of the hierarchy in \eqref{eqn:ineqChain} and of 
the potential usefulness of the bounds. 

%% file: relWork.tex
\subsection{Related Work}
\label{sec:relWork}

Deterministic robustness certification has gained increasing interest in the past few years~\citep{katz2017reluplex,AI2}. Most of the work has focused on verifying properties of the pre-softmax output (e.g.,~\cite{katz2017reluplex,AI2}) for piecewise-linear NNs, while formal reasoning about the softmax outputs themselves has been under-explored. \citet{katz2017reluplex} showed that even the pre-softmax verification problem is computationally inefficient (NP-complete), but a number of approaches based on (linear) abstraction~\citep{singh2019boosting,weng2018towards,zhang2018efficient,gowal2019scalable} and convex optimization~\citep{wong2018provable,dvijotham2018dual} have been proposed to strike a good balance between scalability and precision. We review existing work on NN verification more thoroughly in App.~\ref{sec:nnv-related-work}. 

As mentioned, the softmax function appears in intermediate layers of transformers, and previous works on robustness verification of transformers \citep{Shi2020Robustness,bonaert2021fast} have developed linear lower and upper bounds to approximate the softmax. We review the bounds of \citet{Shi2020Robustness,bonaert2021fast} 
in Section~\ref{sec:ER:lin} as a prelude to deriving provably tighter bounds. \dennis{what else do we want to say about these works?}

Some works have addressed robustness verification of specifications that involve softmax outputs (probabilities) and not just softmax inputs. 
\citet{bitterwolf2020certifiably} obtained an upper bound on the maximal probability (i.e., confidence) to verify the robustness of out-of-distribution detectors and train detectors with such guarantees. Their bound coincides with one of our constant bounds in Section~\ref{sec:prelim:constraints}. \citet{berrada2021make} proposed a general framework for probabilistic specifications on the softmax output, where the NN can be stochastic and inputs can have uncertainty. While the uncertainty quantification metrics that we consider fall under their framework, \citet{berrada2021make} do not give explicit formulations for them, let alone implemented algorithms. \dennis{I re-checked the related work section in \citet{berrada2021make}, the other works seem to be for stochastic NNs/inputs and so less related, but would be good for someone else to check.}

\dennis{other related work in the verification literature?}

Bounds on the softmax and log-sum-exp functions have been used in other contexts. For example, \citet{titsias2016one} derived 
lower bounds on softmax motivated by large-scale classification, \citet{bouchard2007efficient} investigated three upper bounds on log-sum-exp for approximate Bayesian inference, and \citet{nielsen2016guaranteed} used bounds on log-sum-exp to bound information-theoretic measures of mixture models. These bounds, however, do not have the convexity/concavity that we require in the different cases.

%% file: prelim.tex
\section{PRELIMINARIES}
\label{sec:prelim}

For an input $x \in \R^K$, the output $p$ of the softmax function is given by
\begin{equation}\label{eqn:softmax}
    p_j = \frac{e^{x_j}}{\sum_{j'=1}^K e^{x_{j'}}} = \frac{1}{1 + \sum_{j'\neq j} e^{x_{j'} - x_j}}, \quad j = 1,\dots,K.
\end{equation}
We work with the second form above, which is preferred in general for numerical stability and also by \citet{bonaert2021fast} for facilitating their approximations (see their Sec.~5.2). To ease notation, we will focus on the first output $p_1$, without loss of generality because of symmetry. Table~\ref{tab:notation} summarizes the notation used in the paper. Based on the second form in \eqref{eqn:softmax}, we accordingly define $\tx_j \coloneqq x_j - x_1$, $j = 1, \dots, K$. In the simplest case of $K = 2$, the softmax reduces to the 
logistic sigmoid:
\begin{equation}\label{eqn:softmax2}
    p_1 = \frac{1}{1 + e^{\tx_2}} = 1 - p_2.
\end{equation}

We assume that the set of inputs is contained in the hyper-rectangle defined by $l_j \leq x_j \leq u_j$, $j = 1,\dots,K$, which we write as $l \leq x \leq u$. Our goal is to obtain lower bounds $L(x)$ on $p_1$ that are convex functions of $x$, and upper bounds $U(x)$ that are concave functions of $x$, 
\begin{equation}\label{eqn:LU}
L(x) \leq p_1 \leq U(x).
\end{equation}
Constraints of the form in \eqref{eqn:LU} are desirable in general because they define convex sets of $(x, p_1)$ and can be incorporated into convex optimization problems.

\subsection{Basic Bounds and Constraints}
\label{sec:prelim:constraints}

It can be seen from \eqref{eqn:softmax} that $p_1$ is strictly decreasing in $\tx_j = x_j - x_1$ 
for $j \neq 1$. We assume that we have bounds $\tl \leq \tx \leq \tu$ on these differences and also define $\tl_1 = \tu_1 \coloneqq 0$ for the trivial case $\tx_1=0$. Given $l \leq x \leq u$, $\tl_j = l_j - u_1$ and $\tu_j = u_j - l_1$ are always valid bounds on $\tx_j$ for $j \neq 1$, 
but we may have tighter bounds as well. The constraints $\tl \leq \tx \leq \tu$ lead to 
lower and upper bounds on $p_1$:
\begin{subequations}\label{eqn:const}
\begin{align}
    \ulp_1 &= \frac{1}{1 + \sum_{j=2}^K e^{\tu_j}} = \frac{1}{\se(\tu)},\label{eqn:const_LB}\\
    \olp_1 &= \frac{1}{1 + \sum_{j=2}^K e^{\tl_j}} = \frac{1}{\se(\tl)},\label{eqn:const_UB}
\end{align}
\end{subequations}
where we have defined 
the sum-of-exponentials function 
$\se(x) \coloneqq \sum_{j=1}^K e^{x_j}.$ 
These are constant bounds in the sense that they are not functions of $x$. The upper bound \eqref{eqn:const_UB} coincides with the bound of \citet[eq.~(6)]{bitterwolf2020certifiably} given the input bounds $\tl$.

We also have the property that $p$ is non-negative and sums to $1$:
\begin{equation}\label{eqn:simplex}
    \sum_{j=1}^K p_j = 1, \qquad p_j \geq 0 \quad \forall j,
\end{equation}
which are linear constraints on $p$. \citet{bonaert2021fast} recognized the benefit of explicitly enforcing \eqref{eqn:simplex} 
while \citet{Shi2020Robustness} did not use such a constraint.

%% file: ER.tex
\section{BOUNDS FROM EXPONENTIAL\\-RECIPROCAL DECOMPOSITION}
\label{sec:ER}

Previous work \citep{bonaert2021fast} took the natural approach of decomposing the softmax function \eqref{eqn:softmax} as the composition of a sum of exponentials, $q_1 = \se(\tx)$, 
and the reciprocal function $p_1 = 1 / q_1$.\footnote{\citet{bonaert2021fast}'s use of the second form in \eqref{eqn:softmax} avoids a multiplication needed by \citet{Shi2020Robustness}.} They as well as \citet{Shi2020Robustness} derive lower and upper bounds on $\se(\tx)$ and $1/q_1$ that are affine in $x$ and $q_1$ respectively, and compose these bounds to obtain bounds on the softmax. We review the bounds of \citet{Shi2020Robustness,bonaert2021fast} in Section~\ref{sec:ER:lin}, combining their respective advantages, before starting to improve upon them in Section~\ref{sec:ER:non}.

\subsection{Existing Linear Bounds}
\label{sec:ER:lin}

\paragraph{Sum of Exponentials} For the sum of exponentials $\se(\tx)$, each exponential $e^{\tx_j}$ is a function of a scalar $\tx_j \in [\tl_j, \tu_j]$. By virtue of convexity and following \citet{Shi2020Robustness}, each exponential can be bounded from above by the chord between the endpoints $(\tl_j, e^{\tl_j})$ and $(\tu_j, e^{\tu_j})$, and from below by a tangent line passing through $(t_j, e^{t_j})$. The resulting bounds can be written as 
\begin{subequations}\label{eqn:lin_exp}
\begin{align}
    \se(\tx) &\geq 1 + \sum_{j=2}^K e^{t_j} (\tx_j - t_j + 1),\label{eqn:lin_expLB}\\
    \se(\tx) &\leq \olse(\tx; \tl, \tu)
    ,\label{eqn:lin_expUB}
\end{align}
\end{subequations}
where 
\begin{equation}\label{eqn:tj}
t_j = \min\left\{\log\frac{e^{\tu_j} - e^{\tl_j}}{\tu_j - \tl_j}, \tl_j + 1 \right\}
\end{equation}
and we have defined the chordal upper bound on $\se(x)$, 
\begin{equation}\label{eqn:seUB}
    \olse(x; l, u) = \sum_{j=1}^K \left(\frac{u_j - x_j}{u_j - l_j} e^{l_j} + \frac{x_j - l_j}{u_j - l_j} e^{u_j}\right).
\end{equation}
In \eqref{eqn:tj}, the first choice of $t_j$ makes the slope $e^{t_j}$ in the lower bound \eqref{eqn:lin_expLB} equal to the corresponding slope in the upper bound \eqref{eqn:lin_expUB}, thus minimizing the area between them \citep{bonaert2021fast}. The second term in \eqref{eqn:tj} ensures that the lower bound \eqref{eqn:lin_expLB} is non-negative for all $\tx_j \in [\tl_j, \tu_j]$.\footnote{In this second case, we allow the slopes in \eqref{eqn:lin_expLB}, \eqref{eqn:lin_expUB} to be different, like \citet{Shi2020Robustness} and unlike \citet{bonaert2021fast}.} In \eqref{eqn:seUB}, we adopt the convention that if $l_j = x_j = u_j = 0$ (as is true for $\tl_1, \tx_1, \tu_1$), then the $j$th term in the sum is $1$.  

\paragraph{Reciprocal} The same approach is applied to the reciprocal $1/q_1$, which is also a convex function of a scalar. First we need lower and upper bounds on the input $q_1$ to the reciprocal. These are obtained by minimizing the lower bound \eqref{eqn:lin_expLB} and maximizing the upper bound \eqref{eqn:lin_expUB} over $\tx \in [\tl, \tu]$, resulting in
\begin{subequations}\label{eqn:lin_z}
\begin{align}
    \ulq_1^\lin &= 1 + \sum_{j=2}^K e^{t_j} (\tl_j - t_j + 1),\label{eqn:lin_zLB}\\
    \olq_1^\lin &= \se(\tu) = \frac{1}{\ulp_1}.\label{eqn:lin_zUB}
\end{align}
\end{subequations}
Then we have the following bounds on the reciprocal: 
\begin{align}\label{eqn:lin_recip}
    \frac{1}{t_{q_1}} \left(2 - \frac{q_1}{t_{q_1}}\right) \leq \frac{1}{q_1} \leq \frac{1}{\ulq_1^\lin} + \ulp_1 - \frac{\ulp_1 q_1}{\ulq_1^\lin},
\end{align}
where $t_{q_1} = \max\{\sqrt{\ulq_1^\lin \olq_1^\lin}, \olq_1^\lin/2 \}$ is the $q_1$ value of the tangent point.

\paragraph{Softmax} Overall bounds on the softmax output $p_1$ are obtained by composing bounds \eqref{eqn:lin_exp} and \eqref{eqn:lin_recip} with $q_1 = \se(\tx)$ and $p_1 = 1/q_1$. Specifically, upper bound \eqref{eqn:lin_expUB} is composed with the lower bound in \eqref{eqn:lin_recip} to yield the overall lower bound 
\begin{subequations}\label{eqn:lin}
\begin{align}\label{eqn:lin_LB}
    L^\lin(\tx) = &\frac{1}{t_{q_1}} \left(2 - \frac{\olse(\tx; \tl, \tu)}{t_{q_1}} \right). 
\end{align}
Similarly, the combination of \eqref{eqn:lin_expLB} and the upper bound in \eqref{eqn:lin_recip} yield 
\begin{equation}\label{eqn:lin_UB}
    U^\lin(\tx) = \frac{1}{\ulq_1^\lin} + \ulp_1 - \frac{\ulp_1}{\ulq_1^\lin} \left( 1 + \sum_{j=2}^K e^{t_j} (\tx_j - t_j + 1) \right).
\end{equation}
\end{subequations}

\subsection{New Nonlinear Bounds}
\label{sec:ER:non}

We now depart from \citet{Shi2020Robustness,bonaert2021fast} and derive nonlinear bounds on the softmax function using the same exponential-reciprocal decomposition. This is done by further exploiting the convexity of the functions 
$\se(\tx)$ and $1/q_1$. 

\paragraph{Sum of Exponentials} We now regard $q_1$ as an intermediate variable corresponding to the sum of exponentials $\se(\tx)$ but no longer bound by the strict equality $q_1 = \se(\tx)$. For a lower bound on $q_1$, we use $\se(\tx)$ itself, i.e., we relax $q_1 = \se(\tx)$ to 
\begin{equation}\label{eqn:ER_expLB}
    q_1 \geq \se(\tx).
\end{equation}
Since $\se(\tx)$ is convex, the above constraint is in the desired form as in \eqref{eqn:LU}: it specifies a convex set of $(x, q_1)$ and is compatible with convex optimization. On the other hand, for an upper bound on $q_1$, we require a concave function of $x$. We thus reuse the upper bound \eqref{eqn:lin_expUB}, which is linear (and hence concave) in $x$.

\paragraph{Reciprocal} Similarly for the reciprocal where the exact relation is $p_1 = 1/q_1$, we use $1/q_1$ itself as the lower bound on $p_1$ 
and reuse the upper bound in \eqref{eqn:lin_recip}. For the latter however, it is possible to substitute a tighter lower bound on $q_1$ than $\ulq_1^\lin$ in \eqref{eqn:lin_zLB}. The reason is that we can now minimize the lower bound in \eqref{eqn:ER_expLB} over $\tx \in [\tl, \tu]$ instead of the one in \eqref{eqn:lin_expLB}, resulting in 
\begin{equation}\label{eqn:ER_zLB}
    \ulq_1^\ER = \se(\tl) = \frac{1}{\olp_1}. 
\end{equation}
The upper bound on $q_1$ is still $\olq_1^\lin$ \eqref{eqn:lin_zUB}. We therefore have 
\begin{equation}\label{eqn:ER_recip}
    \frac{1}{q_1} \leq p_1 \leq \olp_1 + \ulp_1 - \olp_1 \ulp_1 q_1.
\end{equation}

\paragraph{Softmax} Overall bounds on the softmax are obtained by composing \eqref{eqn:ER_expLB}, \eqref{eqn:lin_expUB} with \eqref{eqn:ER_recip}, specifically the lower bound with the upper bound and vice versa. The results are 
\begin{subequations}\label{eqn:ER}
\begin{align}
    L^\ER(\tx) &= \frac{1}{\olse(\tx; \tl, \tu)}
    ,\label{eqn:ER_LB}\\
    U^\ER(\tx) &= \olp_1 + \ulp_1 - \olp_1 \ulp_1 \se(\tx).\label{eqn:ER_UB}
\end{align}
\end{subequations}
The lower bound $L^\ER(\tx)$ is a composition of $\olse(\tx; \tl, \tu)$, an affine function of $x$, with the reciprocal function. It is thus convex by the composition properties of convex functions \citep[Sec.~3.2.2]{boyd2004convex}. The upper bound $U^\ER(\tx)$ has a sum of exponentials with a negative multiplier in front and is hence concave, as desired.

\begin{theorem}\label{thm:ER_lin}
The nonlinear bounds $L^\ER(\tx)$, $U^\ER(\tx)$ are tighter than the linear bounds $L^\lin(\tx)$, $U^\lin(\tx)$:
\[
L^\lin(\tx) \leq L^\ER(\tx) \leq p_1 \leq U^\ER(\tx) \leq U^\lin(\tx) \quad \forall \tx \in [\tl, \tu].
\]
\end{theorem}
We defer all proofs to Appendix~\ref{sec:proofs}.

%% file: LSE.tex
\section{BOUNDS FROM LOG-SUM-EXP DECOMPOSITION}
\label{sec:LSE}

In this section, we depart from the exponential-reciprocal decomposition altogether and consider an alternative decomposition, obtained by taking the logarithm of the softmax \eqref{eqn:softmax}, $r_1 = -\lse(\tx)$,
and then exponentiating, $p_1 = e^{r_1}$. Here 
$\lse(x) = \log\left( \sum_{j=1}^K e^{x_j} \right)$ 
is the ``log-sum-exp'' (LSE) function, a well-known convex function. Its negative $-\lse(\tx)$ is therefore concave in $x$.

We follow the same approach as in Section~\ref{sec:ER:non}, bounding the exponential and LSE functions and then composing the bounds. 

\paragraph{Exponential} For the exponential function $p_1 = e^{r_1}$, which is again a convex function of a scalar, we use $e^{r_1}$ itself as the lower bound on $p_1$ and a chord of the function as the upper bound. Noting that $\lse(\tx)$ is increasing in all inputs $\tx_j$, $r_1$ is bounded within the interval $[-\lse(\tu), -\lse(\tl)]$, and we may thus use the chord connecting the points $(-\lse(\tu), e^{-\lse(\tu)})$ and $(-\lse(\tl), e^{-\lse(\tl)})$. Using \eqref{eqn:const} to rewrite $-\lse(\tu)$, $-\lse(\tl)$ as $\log(\ulp_1)$, $\log(\olp_1)$, the bounds on $p_1$ in terms of $r_1$ are 
\begin{equation}\label{eqn:LSE_exp}
    e^{r_1} \leq p_1 \leq \frac{\log(\olp_1) - r_1}{\log(\olp_1) - \log(\ulp_1)} \ulp_1 + \frac{r_1 - \log(\ulp_1)}{\log(\olp_1) - \log(\ulp_1)} \olp_1.
\end{equation}

\paragraph{Log-Sum-Exp} For the log-sum-exp function $-\lse(\tx)$, since it is concave in $x$, we may use it as the \emph{upper} bound on its output $r_1$:
\begin{equation}\label{eqn:LSE_lseUB}
    r_1 \leq -\lse(\tx).
\end{equation}

It remains to find a lower bound on $-\lse(\tx)$ that is convex in $x$. In the case $K = 2$, $-\lse(\tx) = -\log(1 + e^{\tx_2})$ is a concave function of a scalar $\tx_2 \in [\tl_2, \tu_2]$ and we may bound it as before using the chord between endpoints,
\begin{multline}\label{eqn:LSE_lseLB2}
    -\log\left(1 + e^{\tx_2}\right) \geq -\frac{\tu_2 - \tx_2}{\tu_2 - \tl_2} \log\left(1 + e^{\tl_2}\right)\\ 
    - \frac{\tx_2 - \tl_2}{\tu_2 - \tl_2} \log\left(1 + e^{\tu_2}\right).
\end{multline}
This is linear and hence convex in $x$.

For $K > 2$, the challenge is that $-\lse(\tx)$ is a multivariate function. Here we provide two bounds with different strengths and weaknesses. In Appendix~\ref{sec:LSE:alt}, we describe a third bound that more directly extends the $K=2$ case \eqref{eqn:LSE_lseLB2} but turns out not to be as tight. For the first bound, we rewrite $-\lse(\tx)$ as $-\lse(\tx) = x_1 - \lse(x)$ so that $-\lse(x)$ is the non-convex part to be bounded. For this, we use chordal bounds on the exponentials similar to \eqref{eqn:lin_expUB}:
\begin{equation}\label{eqn:lseUB}
    \lse(x) 
    \leq \log\left(\olse(x; l, u)\right).
\end{equation}
Hence
\begin{equation}\label{eqn:LSE_lseLB}
    -\lse(\tx) \geq x_1 - \log\left( \olse(x; l, u)\right).
\end{equation}
For the second bound, let $j^* = \argmax_j (l_j + u_j)$ be the index of the largest input in terms of the midpoints $m_j = (l_j + u_j) / 2$. Define the vector of differences $\dot{x}$, $\dot{x}_{j} = x_j - x_{j^*}$, $j = 1,\dots,K$, with corresponding lower and upper bounds $\dot{l}_j$, $\dot{u}_j$. 
We then use the relation $\tx = \dot{x}
- \dot{x}_{1}$ and the translation property of LSE to write $-\lse(\tx) = \dot{x}_{1} - \lse(\dot{x}
)$. By bounding $\lse(\dot{x}
)$ using chordal bounds on exponentials as in \eqref{eqn:lseUB}, we obtain 
\begin{equation}\label{eqn:LSE_lseLB*}
    -\lse(\tx) \geq \dot{x}_{1} - \log\left(\olse(\dot{x}; \dot{l}, \dot{u})\right).
\end{equation}
In both \eqref{eqn:LSE_lseLB} and \eqref{eqn:LSE_lseLB*}, since the term after $x_1$ or $\dot{x}_{1}$ is the composition of an affine function of $x$ with $-\log$, the right-hand sides of \eqref{eqn:LSE_lseLB}, \eqref{eqn:LSE_lseLB*} are convex functions of $x$ as desired. The advantage of \eqref{eqn:LSE_lseLB*} is that the inputs $\dot{x}_{j}$ to $\lse(\dot{x}
)$ tend to be negative, thus placing them in the flatter parts of the exponentials $e^{\dot{x}_{j}}$ and leading to less loss when these exponentials are bounded by chords. The disadvantage of \eqref{eqn:LSE_lseLB*} is that the chords are over intervals $[\dot{l}_{j}, \dot{u}_{j}]$ that tend to be wider than the intervals $[l_j, u_j]$ used in \eqref{eqn:LSE_lseLB}.

\paragraph{Softmax} Overall bounds on the softmax are obtained by composing \eqref{eqn:LSE_lseUB} and \eqref{eqn:LSE_lseLB} or \eqref{eqn:LSE_lseLB*} with \eqref{eqn:LSE_exp}, this time matching lower bound with lower bound and upper with upper:
\begin{subequations}\label{eqn:LSE}
\begin{align}
    L^{\lse}(x) &= \frac{e^{x_1}}{\olse(x; l, u)}
    ,\label{eqn:LSE_LB}\\
    L^{\lse*}(x) &= \frac{e^{\dot{x}_1}}{\olse(\dot{x}; \dot{l}, \dot{u})}
    ,\label{eqn:LSE_LB*}\\
    U^{\lse}(x) &= \frac{\ulp_1 \log(\olp_1) - \olp_1 \log(\ulp_1) - (\olp_1 - \ulp_1) \lse(\tx)}{\log(\olp_1) - \log(\ulp_1)}
    .\label{eqn:LSE_UB}
\end{align}
\end{subequations}
We use the lower bounds in \eqref{eqn:LSE_LB}, \eqref{eqn:LSE_LB*} for $K > 2$ and the upper bound \eqref{eqn:LSE_UB} for all $K$. Bound \eqref{eqn:LSE_LB*} is in fact a generalization of the ER lower bound \eqref{eqn:ER_LB} and coincides with \eqref{eqn:ER_LB} when $j^* = 1$. The lower bounds \eqref{eqn:LSE_LB}, \eqref{eqn:LSE_LB*} are the compositions of the right sides of \eqref{eqn:LSE_lseLB}, \eqref{eqn:LSE_lseLB*}, previously argued to be convex in $x$, with the exponential function, which is convex and increasing. Hence $L^{\lse}(x)$, $L^{\lse*}(x)$ are convex by the composition properties of convex functions \citep[Sec.~3.2.4]{boyd2004convex}. The upper bound $U^{\lse}(x)$ is concave in $x$ as it has $\lse(\tx)$ with a negative multiplier.

For an overall lower bound in the case $K = 2$, we take lower bound \eqref{eqn:LSE_lseLB2} instead of \eqref{eqn:LSE_lseLB} or \eqref{eqn:LSE_lseLB*} and exponentiate. 
After simplifying, this yields
\begin{equation}\label{eqn:LSE_LB2}
    L^{\lse_2}(x) = \left(\ulp_1\right)^{(\tx_2 - \tl_2) / (\tu_2 - \tl_2)} \Bigl(\olp_1\Bigr)^{(\tu_2 - \tx_2) / (\tu_2 - \tl_2)},
\end{equation}
which is an exponential function of $\tx_2$ and hence convex ($\lse_2$ indicates that this bound is only for $K = 2$).

\begin{theorem}\label{thm:LSE_ER}
The log-sum-exp upper bound $U^{\lse}(x)$ is tighter than the nonlinear exponential-reciprocal upper bound $U^\ER(x)$,
\[
p_1 \leq U^{\lse}(x) \leq U^\ER(x) \quad \forall x \in [l, u],
\]
for all $K \geq 2$. The log-sum-exp lower bound $L^{\lse_2}(x)$ is tighter than the nonlinear exponential-reciprocal lower bound $L^\ER(x)$,
\[
L^\ER(x) \leq L^{\lse_2}(x) \leq p_1 \quad \forall x \in [l, u],
\]
for $K=2$.
\end{theorem}
%
%

\begin{figure}[th]
    \centering
    \includegraphics[width=0.85\columnwidth]{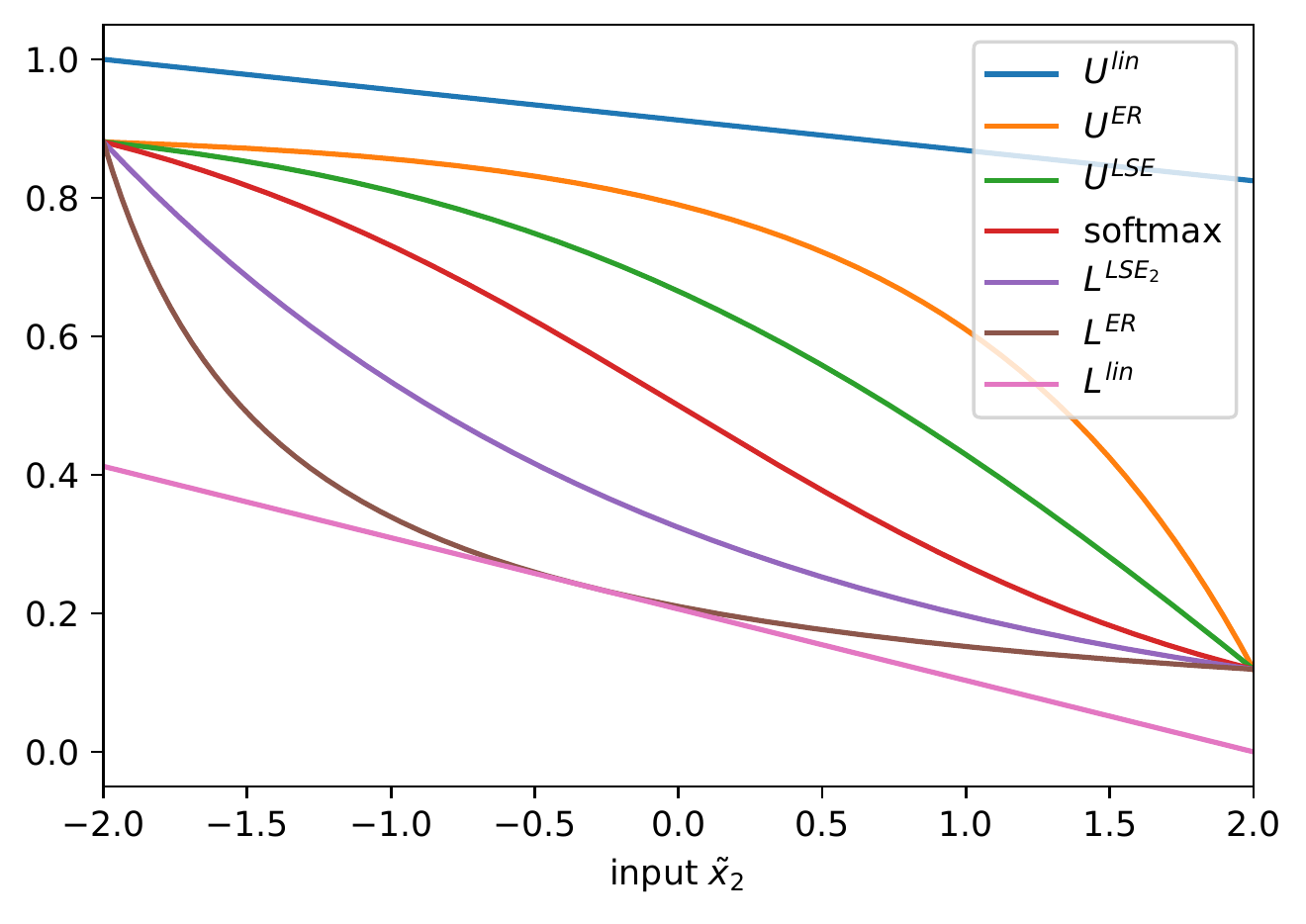}
    \caption{Linear (lin), exponential-reciprocal (ER), and log-sum-exp (LSE) lower and upper bounds on logistic sigmoid function (softmax for $K=2$ inputs).}
    \label{fig:logistic_bounds}
\end{figure}

For $K = 2$ inputs, the softmax function \eqref{eqn:softmax2} and all bounds (linear, ER, LSE) can be plotted as functions of the scalar $\tx_2$. We do so in Figure~\ref{fig:logistic_bounds} for the input interval $[\tl_2, \tu_2] = [-2, 2]$. In addition to confirming Theorems~\ref{thm:ER_lin} and \ref{thm:LSE_ER}, the figure shows that the gap between the ER lower bound $L^\ER$ and softmax is about twice as large as for the LSE lower bound $L^{\lse_2}$, and similarly for the upper bounds. While $L^\lin$ is tangent to $L^\ER$, $U^\lin$ exhibits a larger gap. These observations continue to hold for $K > 2$ in Section~\ref{sec:synth}.

\section{LINEARIZED BOUNDS}
\label{sec:lin}

Any tangent plane to a convex lower bound is also a sound lower bound; and any tangent plane to a concave upper bound is also a sound upper bound. A plane tangent to a function $f:\R^K\mapsto \R$ at point $c$ can be described by:

\begin{equation}
    \overline{f_c}(x) = \sum_{j = 1}^{K} \left(\frac{\partial f(c)}{\partial x_j} \left(x_j- c_j\right)\right) + f(c).
\end{equation}

The coefficient of each addend $(x_j - c_j)$ is the partial derivative of $f(x)$ with respect to $x_j$ evaluated at $c$.  As an example, $\frac{\partial L^{\ER}(x)}{\partial x_j}$ can be computed with the chain rule and the results are:
\begin{subequations}\label{eqn:tan_er_LB}
\begin{align}
\frac{\partial L^{\ER}(x)}{\partial x_1}  &= L^{\ER}(x)^2 \cdot \sum^{K}_{j=2} \left(\frac{e^{\tu_j} - e^{\tl_j}}{\tu_j-\tl_j}\right) \label{eqn:tan_er_LB_x1}\\
\frac{\partial L^{\ER}(x)}{\partial x_j}  &= -L^{\ER}(x)^2 \cdot \frac{e^{\tu_j} - e^{\tl_j}}{\tu_j-\tl_j}  \quad \text{for } j \neq 1. \label{eqn:tan_er_LB_xi}
\end{align}
\end{subequations}
Note that while the above linearized bounds derived from $L^\ER$ and the existing linear bound described in Eq.~\eqref{eqn:lin_LB} are both based on the exponential-reciprocal decomposition, there is a key difference: \eqref{eqn:lin_LB} is obtained by composing the linear over-approximations for each decomposition step, while our approach only linearizes \emph{once} after composing the non-linear bounds. 

Linearized bounds for the other nonlinear 
bounds in Sections~\ref{sec:ER:non} and \ref{sec:LSE} can be found in App.~\ref{sec:other-lin-bounds}.

While the tangent planes as bounds are strictly less tight than their nonlinear counterparts, these linearized bounds can be useful because they can be integrated in existing bound-propagation frameworks~\citep{singh2019abstract,zhang2018efficient} for NNs, which are designed to efficiently bound the outputs of NNs given a set of inputs. We use them to verify self-attention mechanisms in Section~\ref{sec:appl:attention}

%% file: synth.tex
\section{SYNTHETIC DATA EVALUATION}
\label{sec:synth}

\begin{figure*}[t]
\centering
    \begin{subfigure}[b]{0.245\textwidth}
        \centering
        \includegraphics[width=\textwidth]{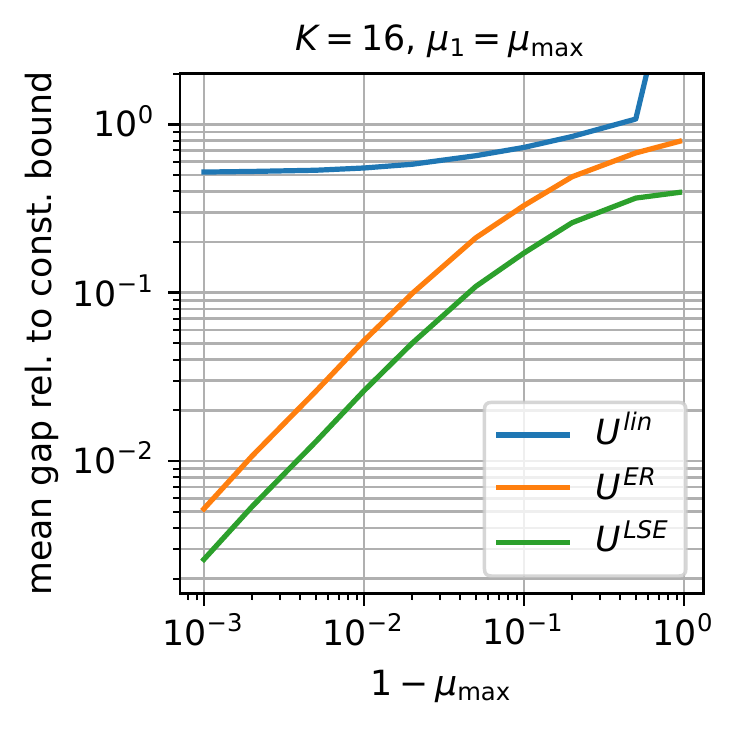}
        \caption{upper bounds, high prob.}
        \label{fig:synth_ubs_rel_d16_high_eps100}
    \end{subfigure}
    \hfill
    \begin{subfigure}[b]{0.245\textwidth}
        \centering
        \includegraphics[width=\textwidth]{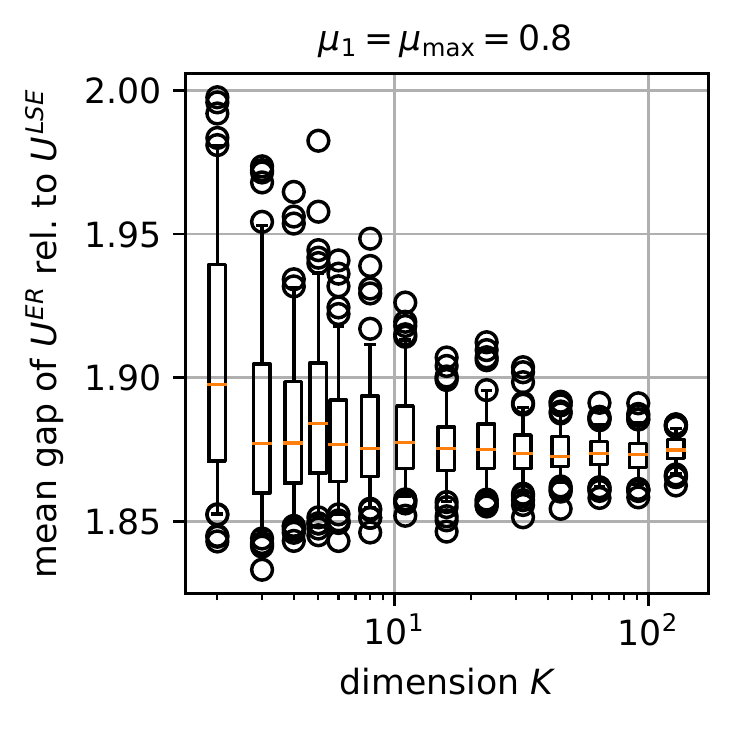}
        \caption{$U^\ER$ vs.~$U^{\lse}$, high prob.}
        \label{fig:synth_ER_u_LSE_u_high0.8_eps100}
    \end{subfigure}
    \hfill
    \begin{subfigure}[b]{0.245\textwidth}
        \centering
        \includegraphics[width=\textwidth]{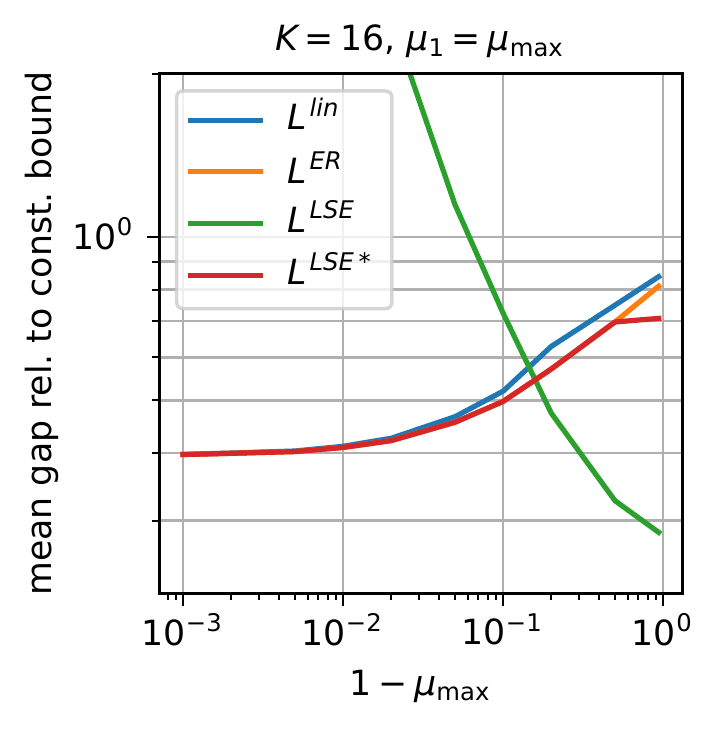}
        \caption{lower bounds, high prob.}
        \label{fig:synth_lbs_rel_d16_high_eps100}
    \end{subfigure}
    \hfill
    \begin{subfigure}[b]{0.245\textwidth}
        \centering
        \includegraphics[width=\textwidth]{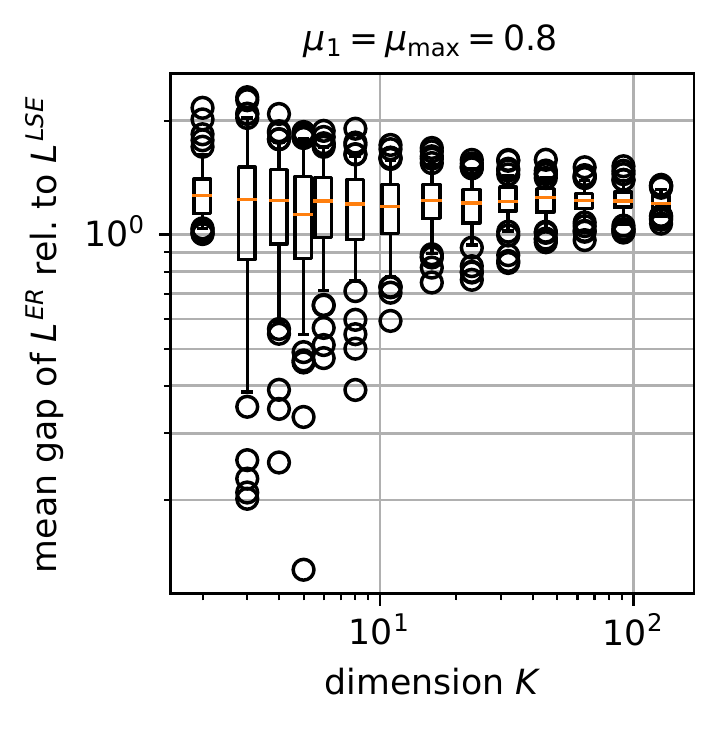}
        \caption{$L^\ER$ vs.~$L^{\lse}$, high prob.}
        \label{fig:synth_ER_l_hybrid_l_high0.8_eps100}
    \end{subfigure}
    \hfill
    \begin{subfigure}[b]{0.245\textwidth}
        \centering
        \includegraphics[width=\textwidth]{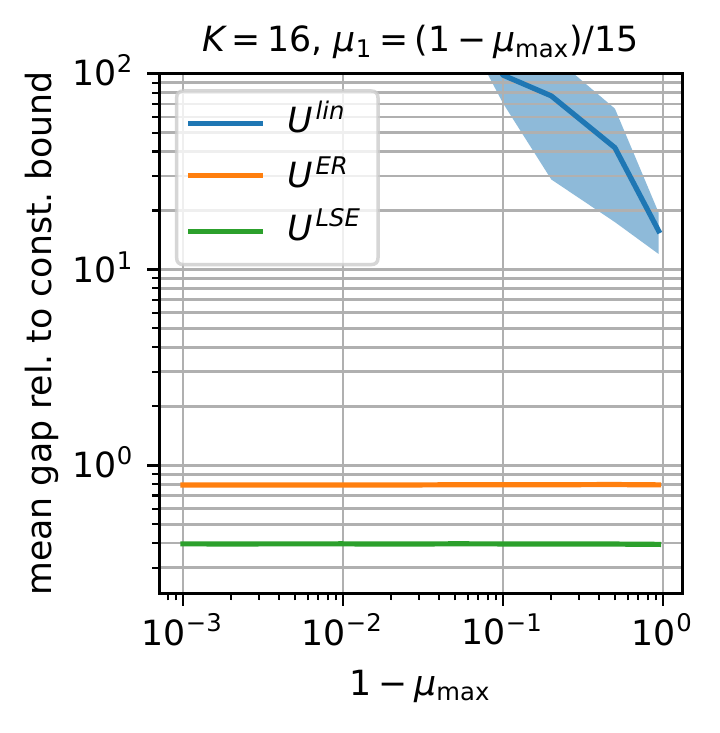}
        \caption{upper bounds, low prob.}
        \label{fig:synth_ubs_rel_d16_low_eps100}
    \end{subfigure}
    \hfill
    \begin{subfigure}[b]{0.245\textwidth}
        \centering
        \includegraphics[width=\textwidth]{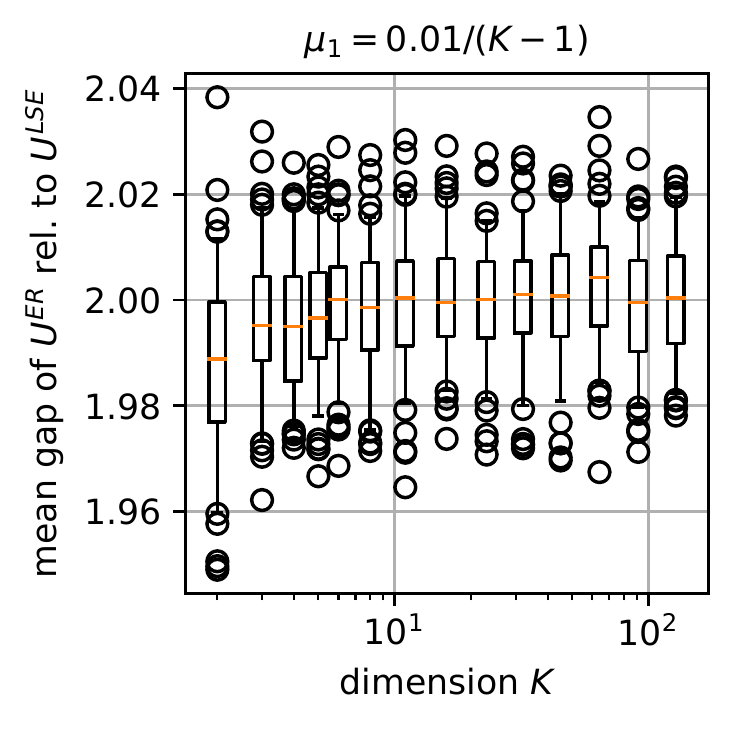}
        \caption{$U^\ER$ vs.~$U^{\lse}$, low prob.}
        \label{fig:synth_ER_u_LSE_u_low0.99_eps100}
    \end{subfigure}
    \hfill
    \begin{subfigure}[b]{0.245\textwidth}
        \centering
        \includegraphics[width=\textwidth]{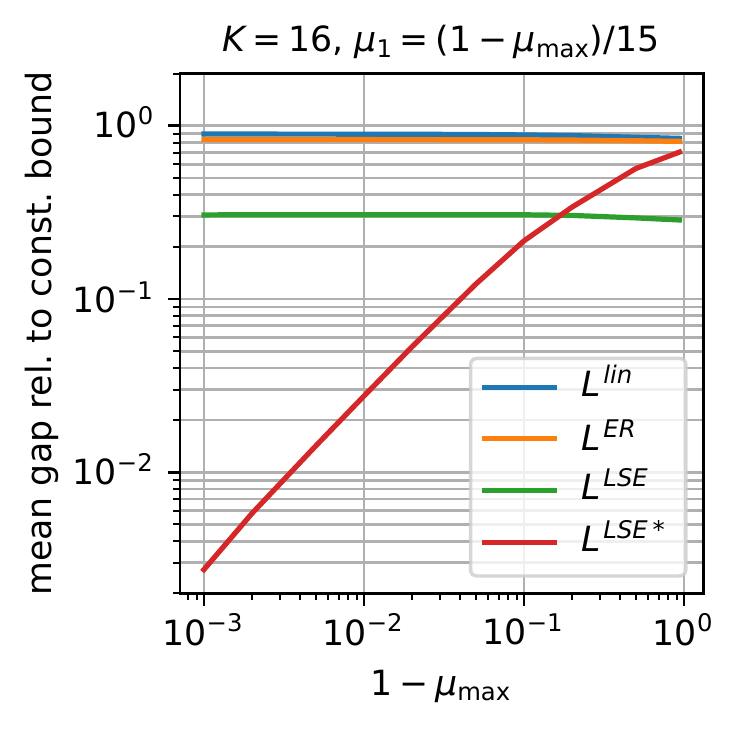}
        \caption{lower bounds, low prob.}
        \label{fig:synth_lbs_rel_d16_low_eps100}
    \end{subfigure}
    \hfill
    \begin{subfigure}[b]{0.245\textwidth}
        \centering
        \includegraphics[width=\textwidth]{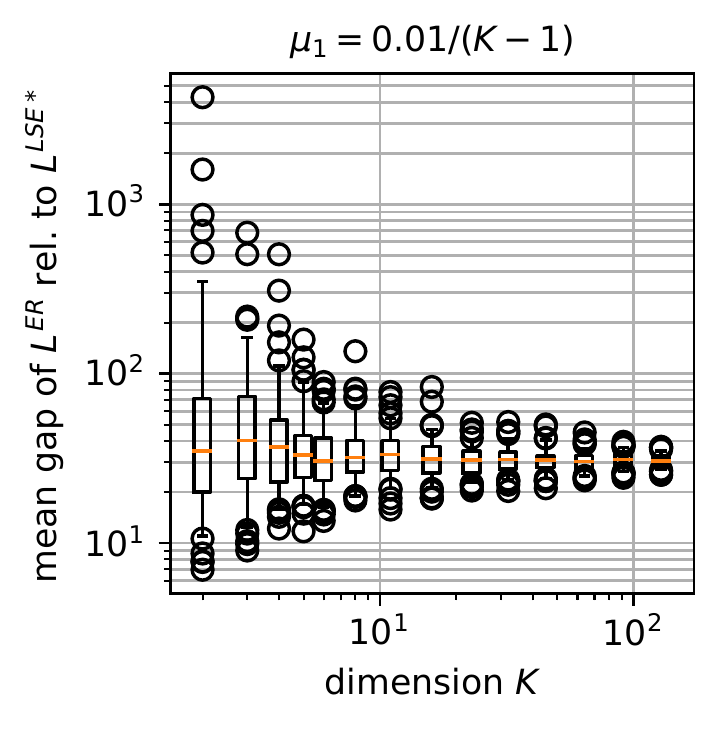}
        \caption{$L^\ER$ vs.~$L^{\lse*}$, low prob.}
        \label{fig:synth_ER_l_hybrid2_l_low0.99_eps100}
    \end{subfigure}
    \hfill
\caption{Mean gaps of upper bounds (left two columns) and lower bounds (right two columns) on softmax output $p_1$ for synthetically generated input regions of width $\epsilon = 1$. In the top/bottom row, the mean $\mu_1$ of $p_1$ is high/low.}
\label{fig:synth}
\end{figure*}

We conduct an experiment using synthetic data to 
compare the tightness of the bounds in Sections~\ref{sec:ER} and \ref{sec:LSE}. For this experiment, we first sample softmax outputs from a $K$-dimensional Dirichlet distribution. To simulate outputs that have varying amounts of probability concentrated on one component, we choose one component of the mean, $\mu_{j_{\max}}$, to be larger than the others, $\mu_{j_{\max}} = \mu_{\max}$, and vary $\mu_{\max}$. More details are in App.~\ref{sec:synth:add}. 
We continue to focus on the first softmax output $p_1$ and consider two cases: $j_{\max} = 1$ ($p_1$ has the largest mean) and $j_{\max} \neq 1$ ($p_1$ is among those with small mean). 
After sampling a softmax output $p$, we convert it to an 
input $m$ (i.e., logits). 
Bounds on the input region are then set as $l_j = m_j - \epsilon$, $u_j = m_j + \epsilon$ for all $j$, where the width $\epsilon$ is varied. Inputs $x$ are sampled from the uniform distribution over the hypercube $[l, u]$. One hundred ($100$) input regions are generated in this manner, and from each region, $1000$ inputs $x$ are sampled.

For each input $x$, we evaluate 
$p_1$ \eqref{eqn:softmax} and the following lower and upper bounds: constant $\ulp_1$, $\olp_1$ \eqref{eqn:const}, linear \eqref{eqn:lin}, ER \eqref{eqn:ER}, and LSE \eqref{eqn:LSE}, \eqref{eqn:LSE_LB2}. We leave the linearized bounds of Section~\ref{sec:lin} to App.~\ref{sec:synth:add}. We compute the \emph{mean gap} $p_1 - L(x)$ between the softmax and each lower bound, where the mean is taken over the uniform samples $x$, and similarly the mean gap $U(x) - p_1$ for each upper bound. 

In Figure~\ref{fig:synth}, we plot ratios of mean gaps to focus more on the comparisons between the various bounds. 
Plots of the mean gaps themselves are in App.~\ref{sec:synth:add}. In Figures~\ref{fig:synth_ubs_rel_d16_high_eps100}, \ref{fig:synth_ubs_rel_d16_low_eps100}, for each input region we divide the mean gap of each upper bound by the mean gap of the constant bound (i.e., $\olp_1 - p_1$). We then plot as a function of $\mu_{\max}$ the mean ratios, taken over the $100$ input regions, as well as the standard errors in the mean. 
Figures~\ref{fig:synth_lbs_rel_d16_high_eps100}, \ref{fig:synth_lbs_rel_d16_low_eps100} are the same for the lower bounds. In Figures~\ref{fig:synth_ER_u_LSE_u_high0.8_eps100}, \ref{fig:synth_ER_u_LSE_u_low0.99_eps100}, we take the ratio of the mean gap of $U^\ER$ to that of $U^{\lse}$ and show box plots over the $100$ input regions (whiskers at the $5$th and $95$th percentiles) for different values of $K$. Figures~\ref{fig:synth_ER_l_hybrid_l_high0.8_eps100}, \ref{fig:synth_ER_l_hybrid2_l_low0.99_eps100} are the same for $L^\ER$ versus $L^{\lse}$ and $L^\ER$ versus $L^{\lse*}$ respectively. The top row of Figure~\ref{fig:synth} represents the case $j_{\max} = 1$, where $p_1$ tends to be high, while the bottom row corresponds to $j_{\max} \neq 1$ and low $p_1$. App.~\ref{sec:synth:add} contains plots for values of $K$ and $\epsilon$ other than those indicated in Figure~\ref{fig:synth}. 

We first discuss the 
upper bounds (left two columns of Figure~\ref{fig:synth}). 
The linear ER bound $U^\lin$ can be quite loose, as previously suggested by Figure~\ref{fig:logistic_bounds} (see App.~\ref{sec:synth:add} for a possible explanation). In Figure~\ref{fig:synth_ubs_rel_d16_low_eps100}, $U^\lin$ is worse than the constant bound $\olp_1$ by at least an order of magnitude. Thus, moving to the novel nonlinear bound $U^\ER$ can already make a big difference. The bound $U^{\lse}$ provides further improvement, as guaranteed by Theorem~\ref{thm:LSE_ER}, and the improvement factor of around $2$ 
is remarkably consistent as a function of $\mu_{\max}$ and $K$. This is particularly evidenced by the narrow distributions of ratios 
in Figures~\ref{fig:synth_ER_u_LSE_u_high0.8_eps100}, \ref{fig:synth_ER_u_LSE_u_low0.99_eps100}.

Turning now to the lower bounds, 
$L^\lin$ is stronger than its counterpart $U^\lin$ in the sense that it improves upon the constant bound $\ulp_1$. The improvement from $L^\lin$ to the novel nonlinear bound 
$L^\ER$ is more marginal. The two LSE bounds $L^{\lse}$ \eqref{eqn:LSE_LB} and $L^{\lse*}$ \eqref{eqn:LSE_LB*} are indeed seen to be complementary as discussed in Section~\ref{sec:LSE}. For $\mu_{\max} \lesssim 0.8$, the largest softmax output does not tend to be that much larger than the others and $L^{\lse}$ is better, whereas for $\mu_{\max} \gtrsim 0.9$, the largest component dominates and $L^{\lse*}$ is better. In the case $j_{\max} \neq 1$ in Figure~\ref{fig:synth_lbs_rel_d16_low_eps100}, the combination of $L^{\lse}$ and $L^{\lse*}$ offer an improvement over $L^\ER$ by a factor ranging from $2.5$--$3$ to much higher. However for $j_{\max} = 1$ and $\mu_{\max} \gtrsim 0.9$ in Figure~\ref{fig:synth_lbs_rel_d16_high_eps100}, $L^{\lse*}$ coincides with $L^\ER$ 
and there is no improvement. In Figures~\ref{fig:synth_ER_l_hybrid_l_high0.8_eps100}, \ref{fig:synth_ER_l_hybrid2_l_low0.99_eps100}, for $K = 2$ (leftmost box plot), we use \eqref{eqn:LSE_LB2} as the LSE lower bound and the box plots confirm the inequality $L^\ER(x) \leq L^{\lse_2}(x)$ from Theorem~\ref{thm:LSE_ER}. For $K > 2$, the median ratio of mean gaps (orange lines) remains approximately constant, although 
a minority of instances have a ratio less than $1$ in Figure~\ref{fig:synth_ER_l_hybrid_l_high0.8_eps100}. 


\dennis{would be best to add the linearized LSE bounds as well}

%% file: appl.tex
\section{APPLICATIONS TO ROBUSTNESS VERIFICATION}
\label{sec:appl}

We present experiments on two robustness verification problems. Our focus remains on showing that the new bounds in Sections~\ref{sec:ER} and \ref{sec:LSE} provide benefits for these tasks, in addition to their theoretical and numerical advantages. 

\subsection{Predictive Uncertainty Estimation}
\label{sec:appl:UQ}

Accurately quantifying uncertainty in predictions is important for calibrating users 
and for identifying highly uncertain and out-of-distribution examples. Many solutions have been proposed for predictive uncertainty estimation with NNs. 
Here we focus on the 
popular technique of using a \emph{deep ensemble} 
of NNs \citep{lakshminarayanan2017simple,rahaman2021uncertainty}. Verification of the robustness of deep ensembles has not been studied to our knowledge. 


To measure the quality of uncertainty estimates, we consider two \emph{proper scoring rules} \citep{gneiting2007strictly}, 
negative log-likelihood (NLL) and Brier score. 
Given an instance $(x^*, y^*)$ with true label $y^*$ ($x$ now refers to the overall NN input) and predicted probabilities $p_k$ 
for 
each class $k = 1, \dots, K$, the scoring rule assigns 
a score $S(p, y^*)$. 
To verify the robustness of uncertainty estimates $p$, we bound 
the worst score that can be attained within an $\ell_p$ ball $\mathcal{B}_p(x^*, \epsilon)$ of radius $\epsilon$ around 
$x^*$. Using the convention that lower scores are better, we thus wish to solve
\begin{equation}\label{eqn:advScore}
    \max_{x \in \mathcal{B}_p(x^*, \epsilon)} S(p, y^*). 
\end{equation}
We formulate \eqref{eqn:advScore} as a concave maximization problem for tractability. Part of this involves expressing $S(p, y^*)$ as, or bounding it from above by, a concave function of $p$. This also suffices for a deep ensemble, where $p$ is the average of probabilities $p^{(m)}$ from the models in the ensemble, since $S(p, y^*)$ or its upper bound will also be concave in $p^{(m)}$.

\paragraph{Negative Log-Likelihood} In the case of NLL, the scoring rule is $S(p, y^*) = -\log p_{y^*}$. While this is not concave in $p$, we can equivalently maximize the linear function 
\begin{equation}\label{eqn:NLL}
    S(p, y^*) = -p_{y^*}.
\end{equation}

\paragraph{Brier Score} Here the scoring rule is 
\begin{align}
    S(p, y^*) = \sum_{k=1}^K \left(p_k - \delta_{k=y^*} \right)^2
    = (1 - p_{y^*})^2 + \sum_{k\neq y^*} p_k^2,
\end{align}
which is a convex sum-of-squares function of $p$. For tractable optimization, we instead maximize an affine upper bound on the Brier score. For each softmax output $p_k$, we have constant bounds $\ulp_k \leq p_k \leq \olp_k$ from \eqref{eqn:const} (generalized to all $k$, and averaged over models for a deep ensemble). We can then bound the convex univariate functions $(1 - p_{y^*})^2$ and $p_k^2$ by the chords connecting their endpoints, as done throughout Sections~\ref{sec:ER} and \ref{sec:LSE}. The resulting bound can be written as the following affine function of $p$:
\begin{equation}\label{eqn:BrierUB}
    S(p, y^*) \leq -2 p_{y^*} + \sum_{k=1}^K (\ulp_{k} + \olp_{k}) p_k - \sum_{k=1}^K \ulp_k \olp_k + 1.
\end{equation}

Given one of the concave scoring objectives in \eqref{eqn:NLL} or \eqref{eqn:BrierUB}, we relate $p$ to the logits predicted by the network(s) using the lower and upper bounds in Sections~\ref{sec:ER} and \ref{sec:LSE}. The logits play the role of $x$ in these bounds. We then encode the remainder of the NN(s), from input to the logits, using existing convex relaxations, and specifically the triangular linear relaxation \citep{ehlers2017formal} in our experiment. For completeness, the full formulation of \eqref{eqn:advScore} as a concave maximization problem is provided in Appendix~\ref{sec:appl:UQ:formulation}.

For our experiment, we train two deep ensembles with $5$ NNs each on the MNIST dataset using the Uncertainty Baselines\footnote{Available from GitHub repository 
\url{https://github.com/google/uncertainty-baselines}} package in Python. 
The architecture of the deep ensemble is given in App.~\ref{sec:arch}. We consider $\ell_\infty$-ball ($\mathcal{B}_\infty(x^*, \epsilon)$) perturbations in \eqref{eqn:advScore}. Input values to the NNs are normalized to between $[0, 1]$. On this scale, we use perturbation bounds $\epsilon = 0.008, 0.012, 0.016$, which correspond to roughly 2, 3, and 4 pixel values. For each of the first $100$ test images, \eqref{eqn:advScore} is solved to bound the score, either NLL or Brier, using different bounds on the softmax. We use CVXPY \citep{diamond2016cvxpy} and its included off-the-shelf solver SCS \citep{scs} to solve these concave problems. We then average over the test images to bound the expected score. 

Table~\ref{tab:UQ} shows the resulting bounds on the expected score. The ``clean'' values are those without perturbation, i.e., $\epsilon = 0$. Recall that the nonlinear bounds denoted $\ER$ and $\lse$ (the two rightmost columns in Table~\ref{tab:UQ}) are both our contributions. While we wished to pair $L^\ER$ with $U^\ER$, using the latter caused the SCS solver to not converge, so we substituted instead the stronger bound $U^{\lse}$. For the LSE pair, we used $L^{\lse*}$ since it suits typical softmax outputs. 
The results indicate that our new nonlinear bounds result in more precise verification of uncertainty quantification than existing linear bounds (represented by $\lin$). In Appendix~\ref{sec:appl:UQ:PGD}, we complement these results with \emph{lower} bounds on worst-case uncertainty estimation scores obtained by using a PGD attack, while in Appendix~\ref{sec:appl:UQ:large}, we show results for an ensemble of larger networks.

\begin{table}[h]
\small
\setlength\tabcolsep{3 pt}
\caption{Upper Bounds on Uncertainty Estimation Scores Using Different Softmax Bounds for the MNIST Classifier} \label{tab:UQ}
\begin{center}
\begin{tabular}{lllll}
\textbf{Score (Clean)}   &$\epsilon$ & $L^\lin, U^\lin$ & $L^\ER, U^{\lse}$ & $L^{\lse*}, U^{\lse}$ \\
\hline 
 NLL (0.105)   & 2/256  & 0.265 & 0.261 & \textbf{0.251} \\
        & 3/256  & 0.442 & 0.433 & \textbf{0.420} \\
        & 4/256  & 0.726 & 0.697 & \textbf{0.690} \\ \hline
Brier (0.048)  & 2/256  & 0.138 & 0.134 & \textbf{0.131} \\
        & 3/256  & 0.244 & 0.235 & \textbf{0.234} \\
        & 4/256  & 0.417 & \textbf{0.403} & \textbf{0.403} \\ \hline
\end{tabular}
\end{center}
\end{table}

We then repeat the same experiment on an ensemble model trained on the CIFAR-10 dataset (architectures can be found in App.~\ref{sec:arch}). Results are shown in Table~\ref{tab:UQ_cifar}.  The results again support the greater strength of the nonlinear bounds over the linear bounds, across $\epsilon$ values and scoring rules. Interestingly, in this case, the $\ER$ lower bound is superior to the $\lse*$ lower bound, suggesting that in practice, $\ER$ and $\lse$ bounds can be complementary. 

\begin{table}[h]
\small
\setlength\tabcolsep{3 pt}
\caption{Upper Bounds on Uncertainty Estimation Scores Using Different Softmax Bounds for the CIFAR-10 Classifier} \label{tab:UQ_cifar}
\begin{center}
\begin{tabular}{lllll}
\textbf{Score (Clean)}   &$\epsilon$ & $L^\lin, U^\lin$ & $L^\ER, U^{\lse}$ & $L^{\lse*}, U^{\lse}$ \\
\hline 
NLL (1.538)   & 2/256  & 2.118 & \bf{2.014} & 2.028 \\
        & 3/256  & 2.569 & \bf{2.433} & 2.474 \\
        & 4/256  & 3.087 & \bf{2.940} & 3.013 \\ \hline
Brier (0.690) & 2/256 & 0.971  & \bf{0.917} & 0.920   \\
        & 3/256  & 1.170 & \bf{1.114} & 1.120  \\
        & 4/256  & 1.367 & \bf{1.324} & 1.329 \\ \hline
\end{tabular}
\end{center}
\end{table}

\paragraph{Timing Results} The average analysis times in seconds per instance using different bounds are shown in Table~\ref{tab:runtime}. The experiments are performed on a cluster equipped with Intel Xeon E5-2637 v4 CPUs. Each job is given one CPU.  

\begin{table}[h]
\caption{Average Runtime in Seconds using Different Bounds}
\label{tab:runtime}
\small
\setlength\tabcolsep{3 pt}
\begin{center}
\begin{tabular}{llll}
\textbf{Dataset} & $L^\lin, U^\lin$ & $L^\ER, U^{\lse}$ & $L^{\lse*}, U^{\lse}$ \\
\hline 
MNIST   & 10.9 & 91.6 & 92.4 \\
CIFAR-10 & 19.5 & 95.3 & 95.5 \\
\hline
\end{tabular}
\end{center}
\end{table}

It must be noted that for this experiment, we are using CVXPY and its off-the-shelf solver SCS, which take time to convert the problems into standard forms and are not customized for them. Hence the runtime results should be interpreted only as confirmation that the convex problems are indeed tractable to solve. In Appendix~\ref{sec:appl:UQ:separate}, we show results from a further over-approximation in which each network in the ensemble is considered separately, which improves computational efficiency.

\subsection{Self-Attention Mechanisms}
\label{sec:appl:attention}

In addition, we consider verifying canonical adversarial robustness properties on neural networks with self-attention mechanisms~\citep{vaswani2017attention}. Self-attention layers involve not only the softmax function, but also bilinear transformations, which are non-trivial to encode as convex optimization problems. Therefore, in this task we instead leverage existing bound-propagation-based methods~\citep{singh2019abstract,zhang2018efficient,Shi2020Robustness}, which already handle bilinear constraints. In particular, we use the CROWN/DeepPoly framework~\citep{singh2019abstract,zhang2018efficient}, a popular bound-propagation method, which requires that each neuron is over-approximated with one linear upper bound and one linear lower bound. Tight convex bounds for bilinear transformation are left as future work. 

\paragraph{MNIST} We train three NNs (named \attSmall, \attMed, and \attBig) with self-attention mechanisms on the MNIST dataset and consider verifying their adversarial robustness against $l_{\infty}$-norm bounded perturbations. The networks are all PGD-trained and vary in size due to different hyper-parameters in the self-attention layers. Details can be found in App.~\ref{sec:arch}. We evaluate on the first 500 test images of the dataset. Inputs to the NNs are again normalized to $[0, 1]$, and we use perturbation bounds 0.016, 0.02, and 0.024, which correspond to roughly 4, 5, and 6 pixel values. 

We consider 4 different linear over-approximations, including the existing linear bounds described in Sec.~\ref{sec:ER:lin} ($\lin$) and 
tangent planes to the non-linear bounds as discussed in Sec.~\ref{sec:lin}. We consider 3 pairs of tangent planes corresponding to the different nonlinear bounds proposed in this work: $\ER$ is derived from $L^{\ER}$ and $U^{\ER}$; $\lse$ is derived from $L^{\lse}$ and $U^{\lse}$; $\lse*$ is derived from $L^{\lse*}$ and $U^{\lse}$. In all cases we take the tangent plane passing through the midpoint of the softmax input range ($\frac{l + u}{2}$).

\begin{table}[h]
\caption{\% of Instances Certified by Different Softmax Bounds for MNIST classifiers} \label{tab:attention}
\begin{center}
\small
\begin{tabular}{llllll}
\textbf{Net. (Acc.) } &\textbf{Pert.} & \textbf{$\lin$} & \textbf{$\ER$} & \textbf{$\lse$} & \textbf{$\lse*$} \\
\hline
        \attSmall (89.2) & 4/256 & 74.0 & 83.6 & 79.4 & \textbf{84.0} \\ 
        ~ & 5/256 & 67.8 & 79.2 & 73.8 & \textbf{81.2} \\ 
        ~ & 6/256 & 61.2 & 74.6 & 68.6 & \textbf{76.4} \\ \hline
        \attMed (98.2) & 4/256 & 60.0 & 81.2 & 84.4 & \textbf{84.6} \\ 
        ~ & 5/256 & 30.0 & 62.4 & 69.2 & \textbf{71.6} \\ 
        ~ & 6/256 & 11.0 & 34.2 & 39.4 & \textbf{46.0} \\ \hline
        \attBig (99) & 4/256 & 42.0 & 65.0 & 68.6 & \textbf{70.8} \\ 
        ~ & 5/256 & 13.0 & 29.4 & 29.6 & \textbf{41.2} \\ 
        ~ & 6/256 & 1.6 & 6.2 & 3.6 & \textbf{11.6} \\ 
        \hline
\end{tabular}
\end{center}
\end{table}

The percentages of verified instances (out of the 500 test images) using different linear bounds are shown in Table~\ref{tab:attention}. Overall, the new linearized bounds (last 3 columns) result in significantly higher verification precision than $\lin$ does. We believe this is because $\lin$ is obtained by linearization at each step of the decomposition, which results in higher accumulation of approximation errors. Among the new linearized bounds, while $\lse*$ consistently certifies more instances than $\ER$, $\lse$ and $\ER$ are evenly matched with head-to-head wins for both. 
Upon further examination, we discover that while $\lse$ is less precise overall, it certifies 9 instances that $\lse*$ is unable to solve and 77 instances that $\ER$ is unable to solve. Overall, the 4 methods combined certify 2628 of the 4500 instances, in contrast to 2572 instances certified by $\lse*$ alone. This suggests the benefit of a portfolio approach. 
Rules for deciding which linearized bounds to use are also an interesting future direction.

\paragraph{SST}
In addition, we perform the same robustness verification task on an NLP transformer model trained on the SST-2 dataset~\citep{socher2013recursive}. SST-2 is a sentiment analysis dataset consisting of movie reviews, where each review is labeled as either positive or negative. In this setting, perturbation is performed on the embedding of the input sentence. The trained transformer obtains 74\% natural accuracy and the robust accuracies verified by different configurations are shown in Table~\ref{tab:attention-sst}. The results further confirm the benefit of the newly proposed linearized bounds (last three columns) over the existing linear bounds ($\lin$). They also highlight the complementary nature of the proposed bounds as unlike in Table~\ref{tab:attention}, $\lse$ is superior to $\lse*$.

\begin{table}[h]
\caption{\% of Instances Certified by Different Softmax Bounds for the SST-2 transformer} \label{tab:attention-sst}
\centering
\small
\begin{tabular}{lllll}
\textbf{Pert. ($\ell_\infty$)} & \textbf{$\lin$} & \textbf{$\ER$} & \textbf{$\lse$} & \textbf{$\lse*$} \\ \hline
0.02 & \textbf{68.8} & \textbf{68.8} & \textbf{68.8} & \textbf{68.8} \\ 
0.04 & \textbf{63.2} & \textbf{63.2} & \textbf{63.2} & \textbf{63.2} \\ 
0.06 & 56.6 & 56.8 & \textbf{57.6} & 57.4 \\ 
0.08 & 51.4 & 52.0 & \textbf{52.4} & \textbf{52.4} \\ 
0.1 & 45.0 & 46.2 & \textbf{47.2} & 46.8 \\ 
0.12 & 39.0 & 39.8 & \textbf{40.6} & \textbf{40.6} \\ 
0.14 & 31.8 & 33.0 & \textbf{34.0} & 33.6 \\ 
0.16 & 26.4 & 26.8 & \textbf{28.2} & 27.2 \\ 
\hline
\end{tabular}
\end{table}

%% file: concl.tex
\section{CONCLUSION}
\label{sec:concl}

We have provided convex bounds on the softmax function satisfying the hierarchy in \eqref{eqn:ineqChain}, both theoretically and numerically, and used them in certifying the robustness of uncertainty estimators and transformers. Future work could consider the development of more customized algorithms for solving the resulting convex optimization problems, and/or further exploitation of the linearized bounds in Section~\ref{sec:lin} to facilitate scaling to larger networks. We also hope for an even better lower bound from the LSE approach, one that might be provably stronger than $L^\ER$ for $K > 2$ or more smoothly integrate the two bounds $L^{\lse}$, $L^{\lse*}$.

\subsubsection*{Artifact}
Scripts to reproduce the experiments in Sections~\ref{sec:synth} and \ref{sec:appl} can be found at \url{https://github.com/NeuralNetworkVerification/bounding-softmax/tree/aistats}.

%% file: appendix.tex
\section{RELATED WORK ON NEURAL NETWORK VERIFICATION \label{sec:nnv-related-work}}

Researchers have proposed several techniques for verifying properties of neural networks~\citep{katz2017reluplex,singh2019abstract,ehlers2017formal,AI2,mipverify,babsr,reluval}. To overcome the inherent scalability limitations, state-of-the-art verifiers seek a good balance between scalability and precision, by designing customized abstractions~\cite{nnv,ehlers2017formal,dlv,deepgame,singh2019abstract,kpoly,star,AI2}, bound-propagation passes~\cite{zelazny2022optimizing,zhang2018efficient,sherlock,mipverify,fastlin,deepz}, or convex optimization procedures~\cite{barrier,barrier-revisited,sdp,bcrown,singh2019boosting,cnn-cert,wu2022efficient}. These abstraction-based methods have been integrated into case-analysis-based search shell to ensure completeness~\citep{katz2017reluplex,marabou,ehlers2017formal,nnenum,nnv,rpm,deepsplit,fromherz2020fast,optAndAbs,mipverify,babsr,peregrinn,dependency,xu2020fast,wu2022efficient}. Our convex optimization procedure could be integrated in a search shell to obtain more precise over-approximation of the output sets.
The linear bounds that we propose can also be integrated into sub-polyhedral abstraction domains~\citep{SinghPV:18} other than DeepPoly/CROWN.

\section{NOTATION}
\label{sec:notation}

Table~\ref{tab:notation} summarizes the more important symbols used in the main paper.

\begin{table}[ht]
    \centering
    \caption{Summary of Notation}
    \label{tab:notation}
    \begin{tabular}{cl}
    \toprule
        \textbf{symbol} & \textbf{description} \\
        \midrule
        $x_j$ & $j$th input to softmax or network \\
        $p_j$ & $j$th output of softmax (probability) \\
        $K$ & number of softmax inputs/outputs \\
        $\tx_j$ & difference $x_j - x_1$ \\
        $\dot{x}_j$ & difference $x_j - x_{j^*}$, where $j^* = \argmax_j (l_j + u_j)$ (defined below) \\
        \midrule
        $l_j$, $u_j$ & lower and upper bounds on $x_j$ \\
        $m_j$ & midpoint $(l_j + u_j) / 2$ of range of $x_j$ \\
        $\epsilon_j$ & half-width $(u_j - l_j ) / 2$ of range of $x_j$ (same for all $j$ if subscript $j$ omitted) \\
        $\tl_j$, $\tu_j$ & lower and upper bounds on $\tx_j$ \\
        $\dot{l}_j$, $\dot{u}_j$ & lower and upper bounds on $\dot{x}_j$ \\
        \midrule
        $\ulp_1$, $\olp_1$ & constant lower and upper bounds on $p_j$ \eqref{eqn:const} \\
        $L_j(x)$, $U_j(x)$ & lower and upper bounds on $p_j$ ($p_1$ if subscript $j$ omitted) as functions of $x$ \\
        $L^\lin(x)$, $U^\lin(x)$ & linear lower and upper bounds on $p_1$ \eqref{eqn:lin} combining \citet{Shi2020Robustness,bonaert2021fast} \\
        $L^\ER(x)$, $U^\ER(x)$ & new nonlinear lower and upper bounds on $p_1$ \eqref{eqn:ER} from exponential-reciprocal (ER) decomposition \\
        $L^{\lse}(x)$ & first nonlinear lower bound on $p_1$ \eqref{eqn:LSE_LB} from log-sum-exp (LSE) decomposition \\
        $L^{\lse*}(x)$ & second nonlinear lower bound on $p_1$ \eqref{eqn:LSE_LB*} from log-sum-exp (LSE) decomposition \\
        $L^{\lse_2}(x)$ & nonlinear lower bound on $p_1$ for $K = 2$ \eqref{eqn:LSE_LB2} from log-sum-exp (LSE) decomposition \\
        $U^{\lse}(x)$ & nonlinear upper bound on $p_1$ \eqref{eqn:LSE_UB} from log-sum-exp (LSE) decomposition \\
        \midrule
        $\se(x)$ & sum-of-exponentials functions $\sum_{j=1}^K e^{x_j}$ \\
        $\olse(x; l, u)$ & chordal upper bound on $\se(x)$ \eqref{eqn:seUB} parametrized by $l, u$ \\
        $\lse(x)$ & log-sum-exp function $\log\left(\sum_{j=1}^K e^{x_j}\right)$ \\
        \midrule
        $q_1$ & intermediate variable in ER decomposition of $p_1$ \\
        $\ulq_1$, $\olq_1$ & constant lower and upper bounds on $q_1$ \eqref{eqn:lin_z} \\
        $t_j$, $t_{q_1}$ & tangent points used in the linear bounds $L^\lin(x)$, $U^\lin(x)$ \\
        $r_1$ & intermediate variable in LSE decomposition of $p_1$ \\
        \midrule
        $x^*$ & clean input \\
        $y^*$ & ground truth label \\
        $S(p, y^*)$ & scoring rule for evaluating $p$ \\
        superscript $m$ & index of model in deep ensemble \\
    \bottomrule
    \end{tabular}
\end{table}

\section{PROOFS}
\label{sec:proofs}

\subsection{Proof of Theorem~\ref{thm:ER_lin}}
\label{sec:ER:proof}

\begin{proof}
The inequality $L^\lin(\tx) \leq L^\ER(\tx)$ is a consequence of applying the first inequality in \eqref{eqn:lin_recip} to \eqref{eqn:lin_LB} (with $q_1 = \olse(\tx; \tl, \tu)$) to yield \eqref{eqn:ER_LB}. 

To establish the inequality $U^\ER(\tx) \leq U^\lin(\tx)$, we first recognize that 
\[
\ulq_1^\lin = 1 + \sum_{j=2}^K e^{t_j} (\tl_j - t_j + 1) \leq \se(\tl) = \frac{1}{\olp_1}
\]
using \eqref{eqn:lin_expLB}. Hence 
\begin{align*}
    U^\ER(\tx) &= \olp_1 \underbrace{\left(1 - \ulp_1 \se(\tx) \right)}_{\geq 0 \text{ for } \tx \in [\tl, \tu]} + \ulp_1\\
    &\leq \frac{1}{\ulq_1^\lin} \left(1 - \ulp_1 \se(\tx) \right) + \ulp_1\\
    &\leq \frac{1}{\ulq_1^\lin} \left(1 - \ulp_1 \left( 1 + \sum_{j=2}^K e^{t_j} (\tx_j - t_j + 1) \right) \right) + \ulp_1\\
    &= U^\lin(\tx),
\end{align*}
where the second inequality is again due to \eqref{eqn:lin_expLB}.
\end{proof}

\subsection{Proof of Theorem~\ref{thm:LSE_ER}}
\label{sec:LSE:proof}

\begin{proof}
To prove the inequality for the \textbf{upper bounds}, we rewrite $U^\ER(x)$ and $U^{\lse}(x)$ as convex combinations of the constant bounds $\ulp_1$ and $\olp_1$. We then compare the coefficients in the two convex combinations.

We rewrite $U^\ER(x)$ \eqref{eqn:ER_UB} as a convex combination as follows, making use of the identities $\ulp_1 \se(\tu) = \olp_1 \se(\tl) = 1$ from \eqref{eqn:const} in the second and fourth lines below:
\begin{align}
    U^\ER(x) &= \olp_1 + \ulp_1 - \olp_1 \ulp_1 \se(\tx) \frac{\se(\tu) - \se(\tl)}{\se(\tu) - \se(\tl)}\nonumber\\
    &= \olp_1 \left(1 - \frac{\se(\tx)}{\se(\tu) - \se(\tl)}\right) + \ulp_1 \left(1 + \frac{\se(\tx)}{\se(\tu) - \se(\tl)}\right)\nonumber\\
    &= \olp_1 \frac{\se(\tu) - \se(\tl) - \se(\tx)}{\se(\tu) - \se(\tl)} + \ulp_1 \frac{\se(\tu) - \se(\tl) + \se(\tx)}{\se(\tu) - \se(\tl)}\nonumber\\
    &= \olp_1 \frac{\se(\tu) - \se(\tx)}{\se(\tu) - \se(\tl)} - \frac{1}{\se(\tu) - \se(\tl)} + \ulp_1 \frac{\se(\tx) - \se(\tl)}{\se(\tu) - \se(\tl)} + \frac{1}{\se(\tu) - \se(\tl)}\nonumber\\
    &= \olp_1 \frac{\se(\tu) - \se(\tx)}{\se(\tu) - \se(\tl)} + \ulp_1 \frac{\se(\tx) - \se(\tl)}{\se(\tu) - \se(\tl)}.\label{eqn:ER_UB_convexCombo}
\end{align}
The two fractions above are non-negative and sum to $1$, so this is indeed a convex combination.

For $U^{\lse}(x)$ \eqref{eqn:LSE_UB}, we use the fact that $-\log(\ulp_1) = \lse(\tu)$ and $-\log(\olp_1) = \lse(\tl)$ to bring it closer to the expression in \eqref{eqn:ER_UB_convexCombo}:
\begin{align}
    U^{\lse}(x) &= \olp_1 \frac{-\log(\ulp_1) - \lse(\tx)}{\log(\olp_1) - \log(\ulp_1)} + \ulp_1 \frac{\lse(\tx) + \log(\olp_1)}{\log(\olp_1) - \log(\ulp_1)}\nonumber\\
    &= \olp_1 \frac{\lse(\tu) - \lse(\tx)}{\lse(\tu) - \lse(\tl)} + \ulp_1 \frac{\lse(\tx) - \lse(\tl)}{\lse(\tu) - \lse(\tl)}.\label{eqn:LSE_UB_convexCombo}
\end{align}
Again the two fractions above are non-negative and sum to $1$.

To show that $U^{\lse}(x) \leq U^\ER(x)$, it now suffices to show that the coefficient of $\olp_1$ in \eqref{eqn:ER_UB_convexCombo} is greater than or equal to the coefficient of $\olp_1$ in \eqref{eqn:LSE_UB_convexCombo} (equivalently, one could compare the $\ulp_1$ coefficients). This can be done using the concavity of the logarithm function as follows:
\begin{align*}
    \lse(\tx) &= \log(\se(\tx))\\
    &\geq \frac{\se(\tu) - \se(\tx)}{\se(\tu) - \se(\tl)} \log(\se(\tl)) + \frac{\se(\tx) - \se(\tl)}{\se(\tu) - \se(\tl)} \log(\se(\tu))\\
    &= \frac{\se(\tu) - \se(\tx)}{\se(\tu) - \se(\tl)} \lse(\tl) + \left(1 - \frac{\se(\tu) - \se(\tx)}{\se(\tu) - \se(\tl)}\right) \lse(\tu).
\end{align*}
The above can be rearranged to yield 
\[
\frac{\se(\tu) - \se(\tx)}{\se(\tu) - \se(\tl)} \geq \frac{\lse(\tu) - \lse(\tx)}{\lse(\tu) - \lse(\tl)},
\]
thus completing the proof for the upper bounds.

To prove the inequality for the \textbf{lower bounds} in the case $K=2$, we rewrite $L^\ER(x)$ \eqref{eqn:ER_LB} as 
\begin{align*}
    L^\ER(x) &= \left( 1 + \frac{\tu_2 - \tx_2}{\tu_2 - \tl_2} e^{\tl_2} + \frac{\tx_2 - \tl_2}{\tu_2 - \tl_2} e^{\tu_2} \right)^{-1}\\
    &= \left( \frac{\tu_2 - \tx_2}{\tu_2 - \tl_2} \left(1 + e^{\tl_2}\right) + \frac{\tx_2 - \tl_2}{\tu_2 - \tl_2} \left(1 + e^{\tu_2}\right) \right)^{-1}\\
    &= \left( \frac{\tu_2 - \tx_2}{\tu_2 - \tl_2} \Bigl(\olp_1\Bigr)^{-1} + \frac{\tx_2 - \tl_2}{\tu_2 - \tl_2} \left(\ulp_1\right)^{-1} \right)^{-1},
\end{align*}
using \eqref{eqn:const} to obtain the last line. This last line can be recognized as a weighted harmonic mean of the constant bounds $\olp_1$ and $\ulp_1$, with weights $(\tu_2 - \tx_2) / (\tu_2 - \tl_2)$ and $(\tx_2 - \tl_2) / (\tu_2 - \tl_2)$. On the other hand, for $K = 2$, $L^{\lse_2}(x)$ \eqref{eqn:LSE_LB2} is a weighted geometric mean of the same quantities with the same weights. It follows from the inequality of (weighted) harmonic and geometric means\footnote{This can be proven as a corollary of the arithmetic mean-geometric mean inequality, among other ways.} that $L^\ER(x) \geq L^{\lse_2}(x).$ 
\end{proof}

\section{ALTERNATIVE LOG-SUM-EXP LOWER BOUND FOR $K > 2$}
\label{sec:LSE:alt}

This appendix describes a third lower bound arising from the log-sum-exp decomposition of Section~\ref{sec:LSE}, as an alternative to \eqref{eqn:LSE_lseLB}, \eqref{eqn:LSE_lseLB*}.

Our motivation is to generalize inequality \eqref{eqn:LSE_lseLB2}, which applies when $K = 2$ and leads to a provably tighter overall bound \eqref{eqn:LSE_LB2} than  $L^\ER(x)$ \eqref{eqn:ER_LB} (Theorem~\ref{thm:LSE_ER}). We start by rewriting $-\lse(\tx)$ as 
\begin{equation}\label{eqn:lse2}
    -\lse(\tx) = -\log\left(1 + e^{-x_1} \sum_{j=2}^K e^{x_j} \right) = -\log\left(1 + e^{\lse(x_2^K) - x_1} \right),
\end{equation}
where $x_2^K = (x_2, \dots, x_K)$. We then apply inequality \eqref{eqn:LSE_lseLB2} to \eqref{eqn:lse2} with $\lse(x_2^K) - x_1$ in place of $\tx_2$. To do this, we also have to replace $\tl_2$, $\tu_2$ with lower and upper bounds on $\lse(x_2^K) - x_1$. Given the monotonicity of $\lse(x_2^K)$, we use the bounds 
\begin{subequations}
\begin{align}
    \ulv_1 &= \lse\left(l_2^K\right) - u_1,\\
    \olv_1 &= \lse\left(u_2^K\right) - l_1.
\end{align}
\end{subequations}
Then the application of \eqref{eqn:LSE_lseLB2} to \eqref{eqn:lse2} yields 
\begin{equation}\label{eqn:LSE_lseLB0}
    -\lse(\tx) \geq \underbrace{-\frac{\log(\olp_1) - \log(\ulp_1)}{\olv_1 - \ulv_1}}_{\leq 0} \left(\lse\left(x_2^K\right) - x_1\right)
    + \frac{\olv_1 \log(\olp_1) - \ulv_1 \log(\ulp_1)}{\olv_1 - \ulv_1},
\end{equation}
where \eqref{eqn:const} has been used to identify $-\log(1 + e^{\ulv_1}) = \log(\olp_1)$ and $-\log(1 + e^{\olv_1}) = \log(\ulp_1)$.

For $K = 2$, the lower bound in \eqref{eqn:LSE_lseLB0} is affine in $x$ and hence convex, but for $K > 2$, it is concave because of the negative multiplier in front. To address the non-convexity, we further bound $\lse(x_2^K)$ using chordal bounds similar to \eqref{eqn:lin_expUB}:
\begin{equation}\label{eqn:lse2UB}
    \lse\left(x_2^K\right) = \log\left( \sum_{j=2}^K e^{x_j} \right)
    \leq \log\left( \sum_{j=2}^K \left(\frac{u_j - x_j}{u_j - l_j} e^{l_j} + \frac{x_j - l_j}{u_j - l_j} e^{u_j}\right) \right).
\end{equation}
Substituting \eqref{eqn:lse2UB} into \eqref{eqn:LSE_lseLB0} gives 
\begin{multline}\label{eqn:LSE_lseLBalt}
    -\lse(\tx) \geq \frac{\olv_1 \log(\olp_1) - \ulv_1 \log(\ulp_1)}{\olv_1 - \ulv_1} + \frac{\log(\olp_1) - \log(\ulp_1)}{\olv_1 - \ulv_1}
    \left(x_1 
    - \log\left( \sum_{j=2}^K \left(\frac{u_j - x_j}{u_j - l_j} e^{l_j} + \frac{x_j - l_j}{u_j - l_j} e^{u_j}\right) \right) \right).
\end{multline}
Since the term after $x_1$ is the composition of an affine function of $x_2^K$ with $-\log$, the right-hand side of \eqref{eqn:LSE_lseLBalt} is now a convex function of $x$.

An overall lower bound on the softmax function is obtained by exponentiating \eqref{eqn:LSE_lseLBalt}:
\begin{align}
    L^{\lse'}(x) &= \exp\left[ \frac{\olv_1 \log(\olp_1) - \ulv_1 \log(\ulp_1)}{\olv_1 - \ulv_1} + \frac{\log(\olp_1) - \log(\ulp_1)}{\olv_1 - \ulv_1} 
    \left(x_1 - \log\left( \sum_{j=2}^K \left(\frac{u_j - x_j}{u_j - l_j} e^{l_j} + \frac{x_j - l_j}{u_j - l_j} e^{u_j}\right) \right) \right) \right].\label{eqn:LSE_LBalt}
\end{align}
In the additional synthetic experiment results reported in Appendix~\ref{sec:synth:add}, we do not see a regime in which $L^{\lse'}$ is better than the larger of $L^{\lse}$, $L^{\lse*}$, i.e., no regime in which $L^{\lse'}$ is the uniquely best lower bound. We did not include $L^{\lse'}$ in the main paper for this reason.

\section{LINEARIZED BOUNDS \label{sec:other-lin-bounds}}

We here present the partial derivatives of the non-linear lower- and upper- bounds presented in the paper, in addition to those of $L^{\ER}$.

$\frac{\partial L^{\lse}(x)}{\partial x_i}$ can be computed by applying the product rule and the chain rule. The results are:
\begin{align}
\frac{\partial L^{\lse}(x)}{\partial x_1}  &= L^{\lse}(x) - e^{x_1} s(x)^2 \left(\frac{e^{u_i} - e^{l_i}}{u_i - l_i}\right) \label{eqn:tan_lse1_LB_x1}\\
\frac{\partial L^{\lse}(x)}{\partial x_i}  &=  - e^{x_1} s(x)^2 \left(\frac{e^{u_i} - e^{l_i}}{u_i - l_i}\right)\label{eqn:tan_lse1_LB_xi} \quad \text{for } i \neq 1
\end{align}

where $s(x) = 1 / \olse(x; l, u)$ 
from \eqref{eqn:LSE_LB}.

$\frac{\partial L^{\lse*}(x)}{\partial x_i}$ is slightly more complicated. When $j^* = 1$, $\frac{\partial L^{\lse*}(x)}{\partial x_i}$ is the same as $\frac{\partial L^{\ER}(x)}{\partial x_i}$. In the case where $j^* \neq 1$, $\frac{\partial L^{\lse*}(x)}{\partial x_i}$ can be again computed with applications of chain rules. The results are:
\begin{align}
\frac{\partial L^{\lse*}(x)}{\partial x_1}  &= 
L^{\lse*}(x)
- \frac{e^{\dot{x}_1}}{\olse(\dot{x}, \dot{l}, \dot{u})^2} 
\frac{e^{\dot{u}_1}-e^{\dot{l}_1}}{\dot{u}_1-\dot{l}_1}
\label{eqn:tan_lse2_LB_x1}\\
\frac{\partial L^{\lse*}(x)}{\partial x_{j^*}}  &= -L^{\lse*}(x) + \frac{e^{\dot{x}_1}}{\olse(\dot{x}, \dot{l}, \dot{u})^2} 
\cdot \sum^{K}_{i\neq j^*} \left(\frac{e^{\dot{u}_i} - e^{\dot{l}_i}}{\dot{u}_i- \dot{l}_i}\right)
\label{eqn:tan_lse2_LB_x*}\\
\frac{\partial L^{\lse*}(x)}{\partial x_i}  &= 
- \frac{e^{\dot{x}_1}}{\olse(\dot{x}, \dot{l}, \dot{u})^2} 
\frac{e^{\dot{u}_i}-e^{\dot{l}_i}}{\dot{u}_i-\dot{l}_i}
\label{eqn:tan_lse2_LB_xi} \quad \text{for } i \not\in \{1,j^*\}
\end{align}


$\frac{\partial U^{\ER}(x)}{\partial x_i}$ can be computed with the chain rule. The results are:
\begin{align}
\frac{\partial U^{\ER}(x)}{\partial x_1}  &= \olp_1 \ulp_1 \left(\se(\tx) - 1\right) \label{eqn:tan_er_UB_x1}\\
\frac{\partial U^{\ER}(x)}{\partial x_i}  &= -\olp_1 \ulp_1 e^{\tx_i} \label{eqn:tan_er_UB_xi} \quad \text{for } i \neq 1
\end{align}

The partial derivative $\frac{\partial U^{\lse}(x)}{\partial x_i}$ can be computed by rewriting $\lse(\tx)$ as $\lse(x) - x_1$, and applying the chain rule and the fact that $\frac{\partial\lse(x)}{\partial x_i} = \frac{e^{x_i}}{\se(x)}$. The results are:
\begin{align}
\frac{\partial U^{\lse}(x)}{\partial x_1}  &= - \frac{\olp_1 - \ulp_1}{\log(\olp_1) - \log(\ulp_1)} \left(\frac{e^{x_1}}{\se(x)} - 1\right) \\
\frac{\partial U^{\lse}(x)}{\partial x_i}  &= - \frac{\olp_1 - \ulp_1}{\log(\olp_1) - \log(\ulp_1)} \cdot \frac{e^{x_i}}{\se(x)} \quad \text{for } i \neq 1
\end{align}

\section{SYNTHETIC DATA EVALUATION: DETAILS AND VARIATIONS}
\label{sec:synth:add}

This appendix contains additional material on the synthetic experiment in Section~\ref{sec:synth}: details on data generation, variations of Figure~\ref{fig:synth} in the main paper, and an explanation of the looseness of the $U^\lin$ bound. 

\subsection{Data Generation Details} Recall that we sample softmax outputs from a Dirichlet distribution and choose one component of the Dirichlet mean, $\mu_{j_{\max}}$, to be larger than the others, $\mu_{j_{\max}} = \mu_{\max} \geq \mu_j$ for $j \neq j_{\max}$. This is done by setting the Dirichlet concentration parameters to be $\alpha_{j_{\max}} = \alpha_{\max} \geq 1$ and $\alpha_j = 1$ for $j \neq j_{\max}$. Then the largest mean component is given by 
\[
\mu_{\max} = \frac{\alpha_{\max}}{\sum_{j=1}^K \alpha_j} = \frac{\alpha_{\max}}{\alpha_{\max} + K - 1}.
\]
After sampling a softmax output $p$, we convert it to an 
input $m$ (i.e., logits) by taking $m_j = \log(p_j / p_1)$ and then centering $m$ by subtracting the mean of the $m_j$'s. 

\begin{figure*}[t!]
\centering
    \begin{subfigure}[b]{0.4\textwidth}
        \centering
        \includegraphics[width=\textwidth]{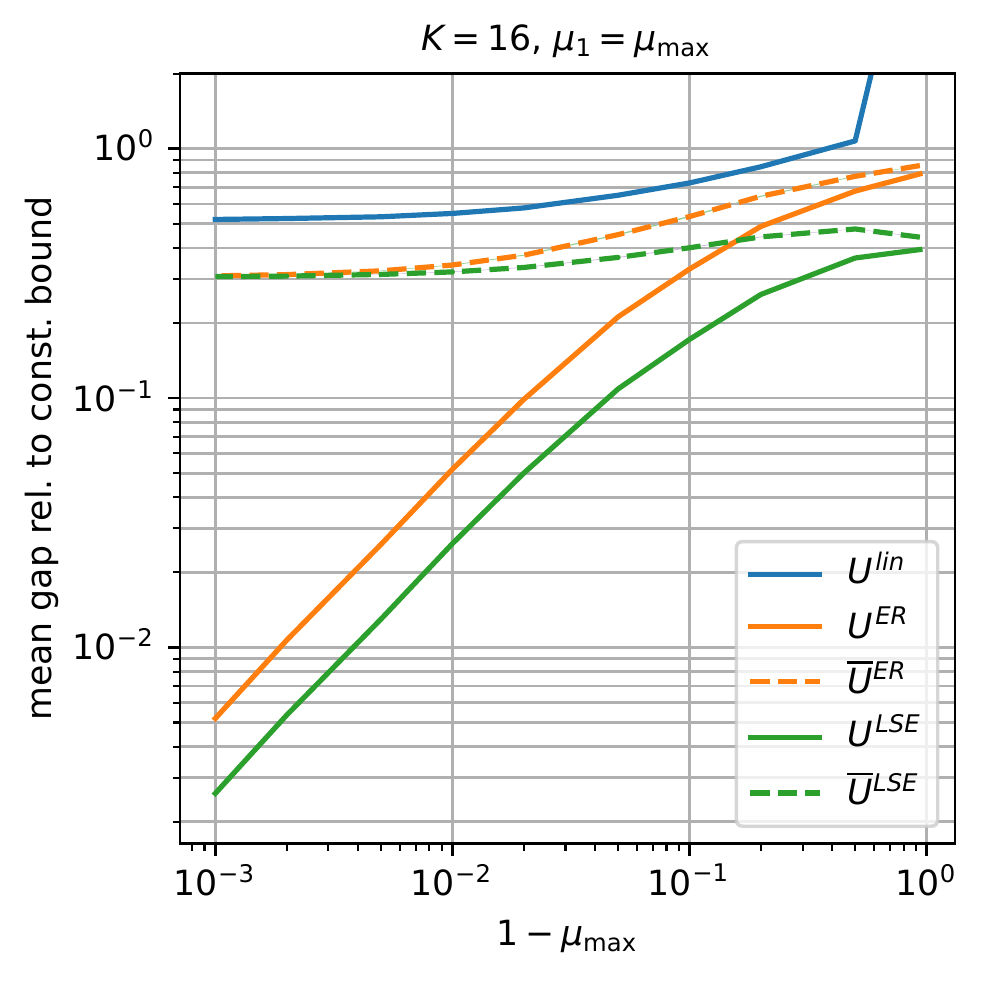}
        \caption{upper bounds, high prob.}
        \label{fig:synth_ubs_lin_d16_high_eps100}
    \end{subfigure}
    \begin{subfigure}[b]{0.4\textwidth}
        \centering
        \includegraphics[width=\textwidth]{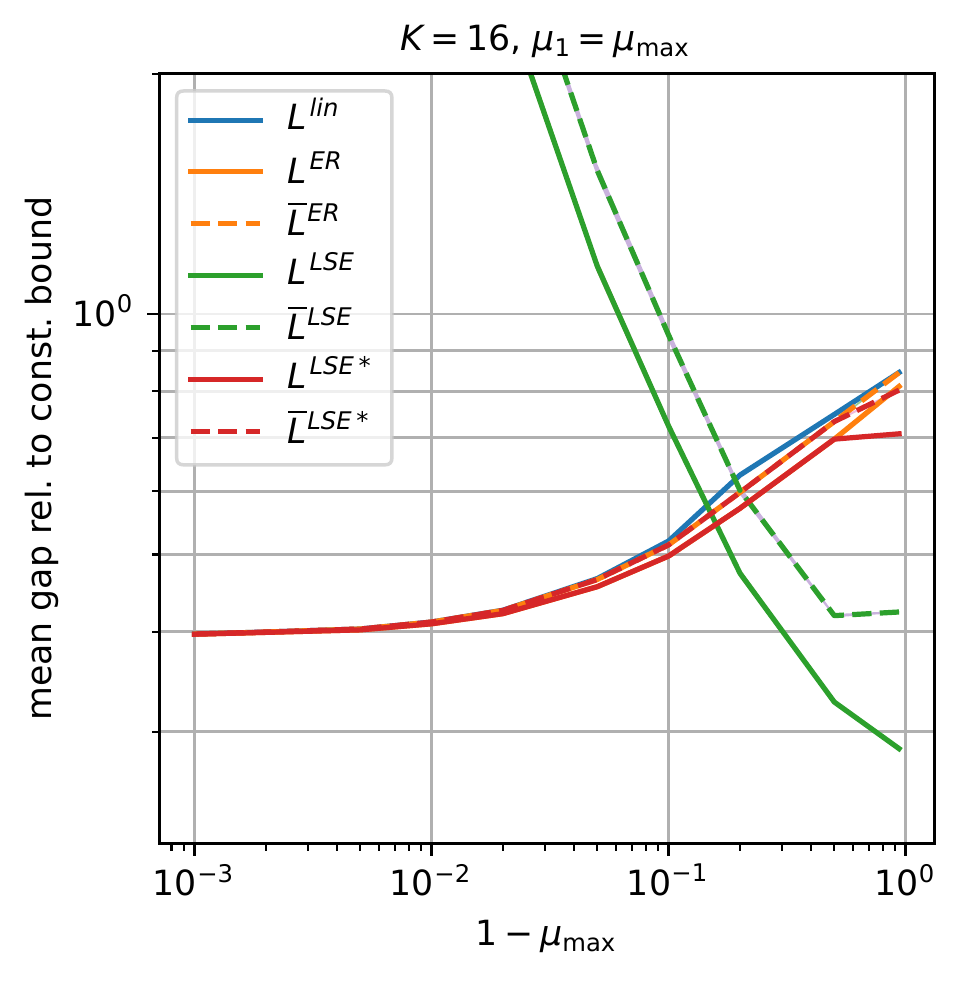}
        \caption{lower bounds, high prob.}
        \label{fig:synth_lbs_lin_d16_high_eps100}
    \end{subfigure}
    \hfill
    \begin{subfigure}[b]{0.4\textwidth}
        \centering
        \includegraphics[width=\textwidth]{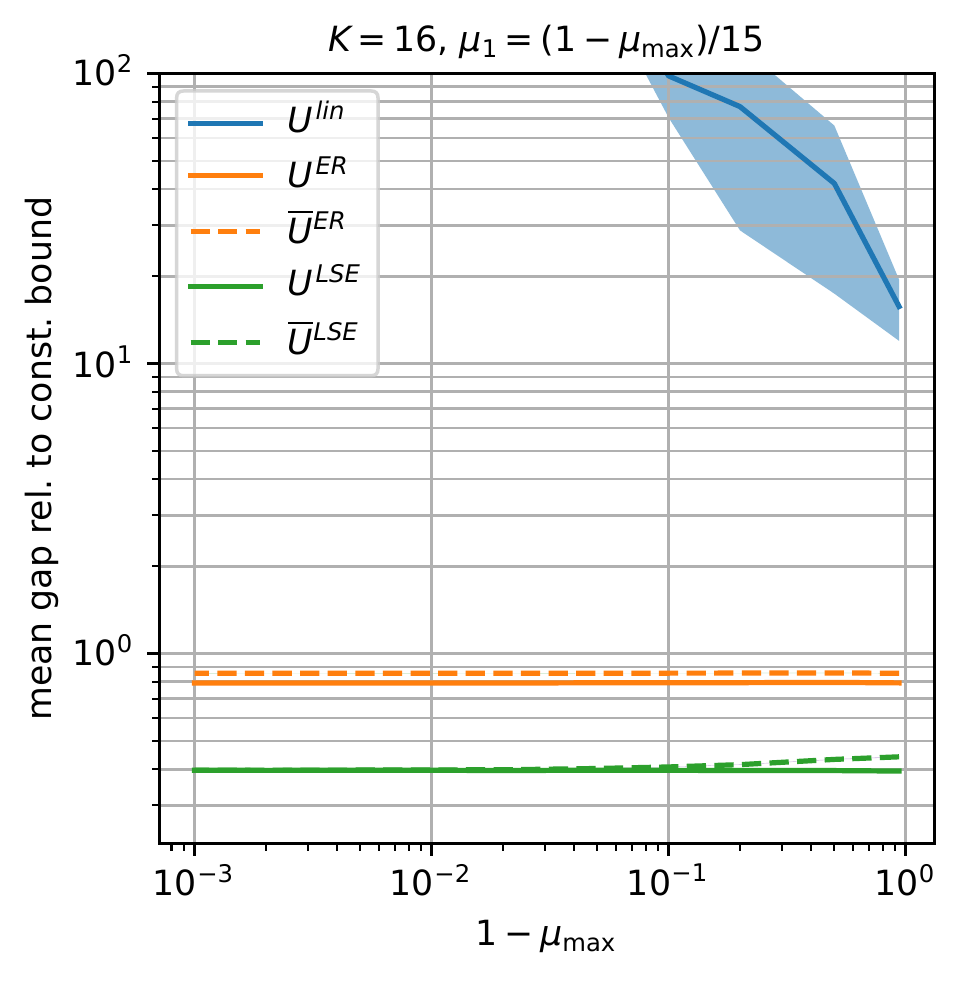}
        \caption{upper bounds, low prob.}
        \label{fig:synth_ubs_lin_d16_low_eps100}
    \end{subfigure}
    \begin{subfigure}[b]{0.4\textwidth}
        \centering
        \includegraphics[width=\textwidth]{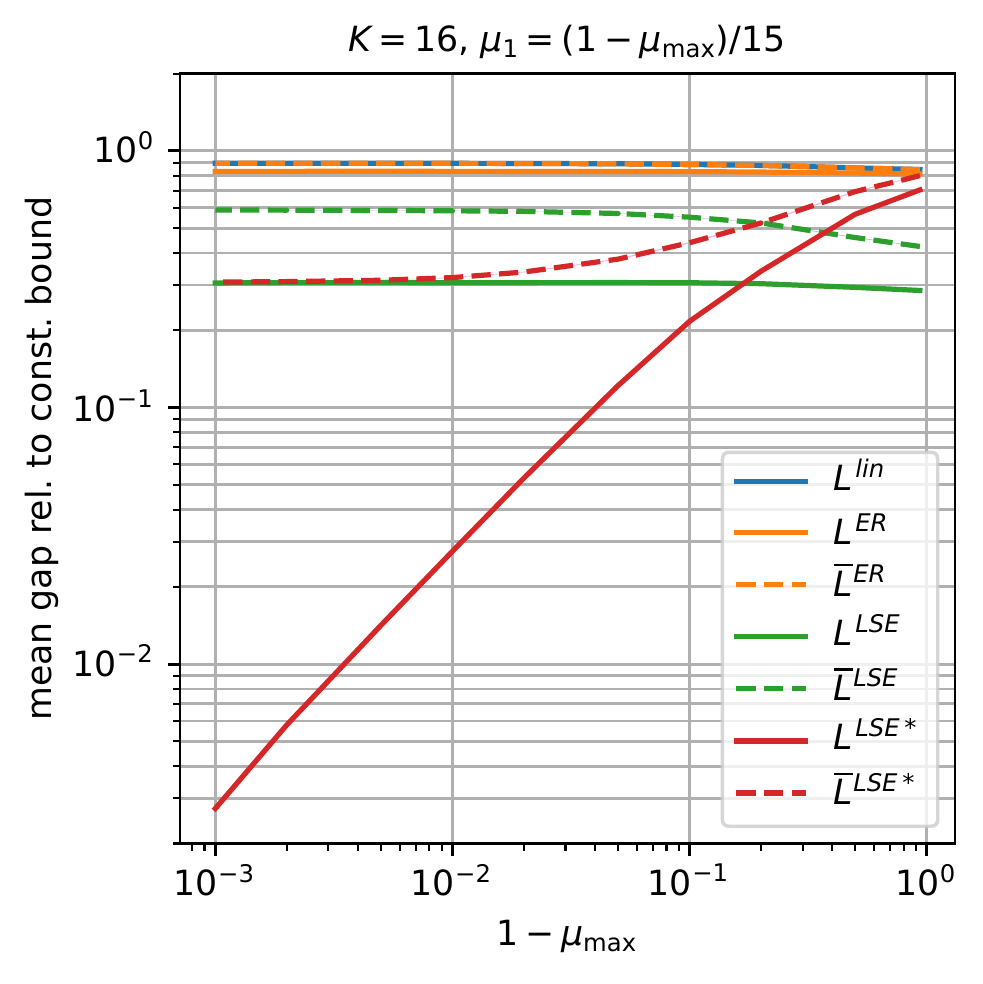}
        \caption{lower bounds, low prob.}
        \label{fig:synth_lbs_lin_d16_low_eps100}
    \end{subfigure}
\caption{Mean gaps of upper bounds (left column) and lower bounds (right column) on softmax output $p_1$, now including linearized bounds (dashed lines), for synthetically generated input regions of width $\epsilon = 1$. In the top/bottom row, the mean $\mu_1$ of $p_1$ is high/low.}
\label{fig:synth_lin}
\end{figure*}

\subsection{Variations on Figure~\ref{fig:synth}}

\paragraph{Linearized Bounds} In Figure~\ref{fig:synth_lin}, we add the linearized bounds of Section~\ref{sec:lin} to Figures~\ref{fig:synth_ubs_rel_d16_high_eps100}, \ref{fig:synth_lbs_rel_d16_high_eps100}, \ref{fig:synth_ubs_rel_d16_low_eps100}, \ref{fig:synth_lbs_rel_d16_low_eps100}. The linearized bounds are plotted using dashed lines of the same color as their non-linearized counterparts. In Figures~\ref{fig:synth_lbs_lin_d16_high_eps100}, \ref{fig:synth_ubs_lin_d16_low_eps100} (lower bounds on a probability with high mean and upper bounds on a low probability), the losses in strength due to linearization are modest. However in Figures~\ref{fig:synth_ubs_lin_d16_high_eps100}, \ref{fig:synth_lbs_lin_d16_low_eps100}, the mean gap ratios of the linearized bounds (relative to the respective constant bound) appear to be limited to no lower than $0.3$, whereas the mean gap ratios of $U^\ER$, $U^{\lse}$, and $L^{\lse*}$ decrease to much smaller values. Thus the gain due to nonlinearity appears to be substantial in these two scenarios (upper bound on high probability and lower bound on low probability).

\clearpage

\begin{figure*}[th]
\centering
    \begin{subfigure}[b]{0.4\textwidth}
        \centering
        \includegraphics[width=\textwidth]{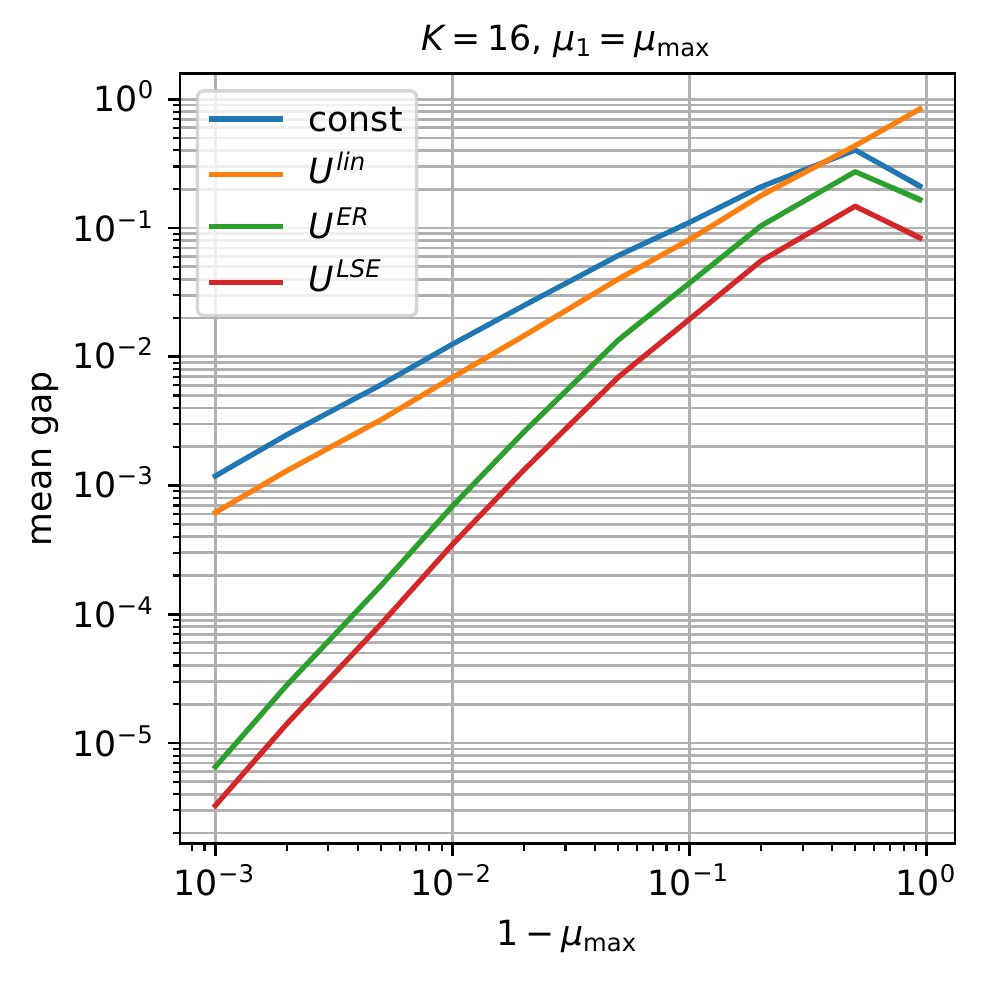}
        \caption{upper bounds, high prob.}
        \label{fig:synth_ubs_d16_high_eps100}
    \end{subfigure}
    \begin{subfigure}[b]{0.4\textwidth}
        \centering
        \includegraphics[width=\textwidth]{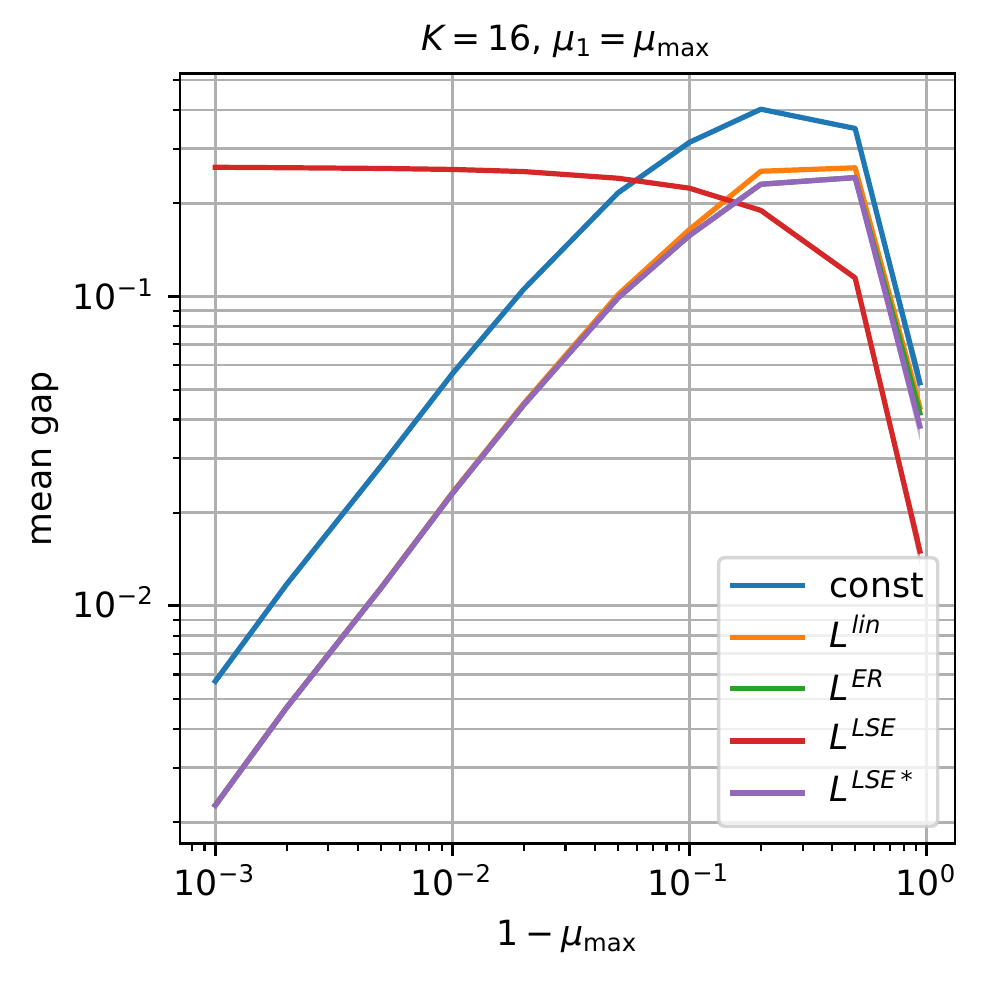}
        \caption{lower bounds, high prob.}
        \label{fig:synth_lbs_d16_high_eps100}
    \end{subfigure}
    \hfill
    \begin{subfigure}[b]{0.4\textwidth}
        \centering
        \includegraphics[width=\textwidth]{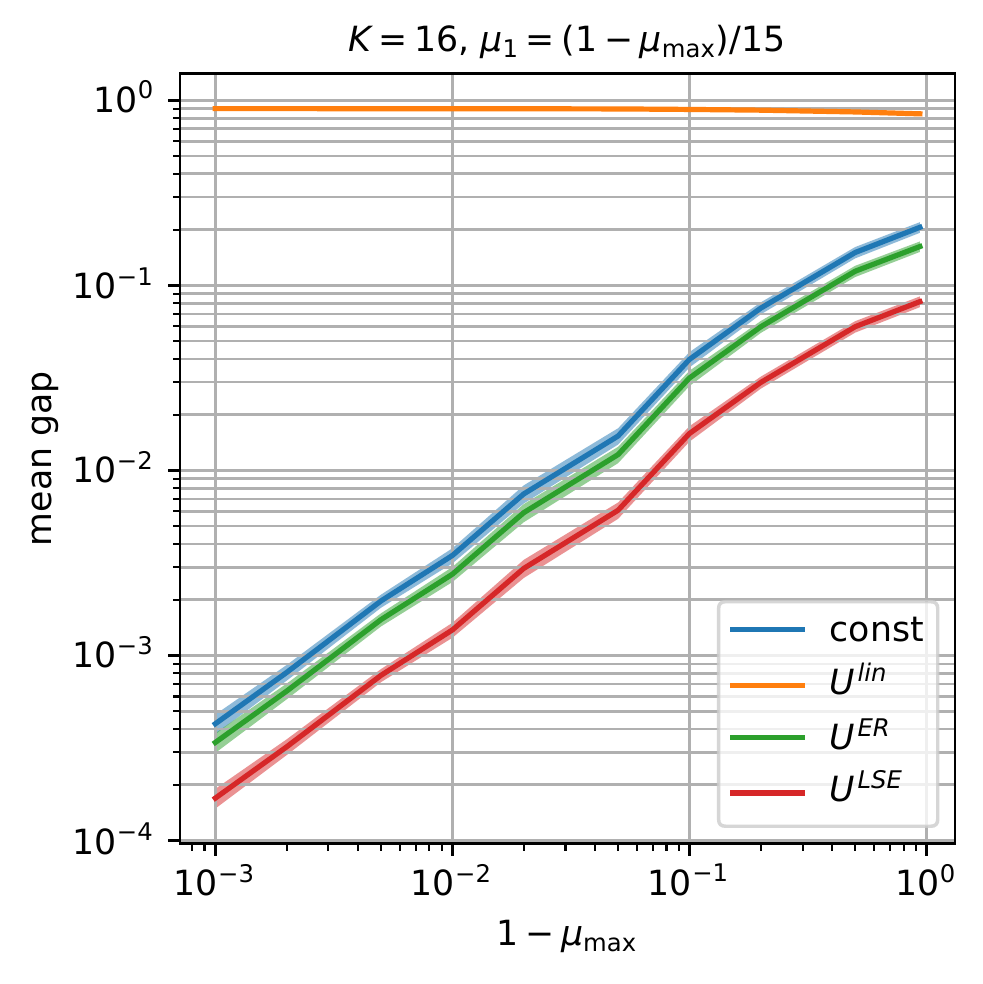}
        \caption{upper bounds, low prob.}
        \label{fig:synth_ubs_d16_low_eps100}
    \end{subfigure}
    \begin{subfigure}[b]{0.4\textwidth}
        \centering
        \includegraphics[width=\textwidth]{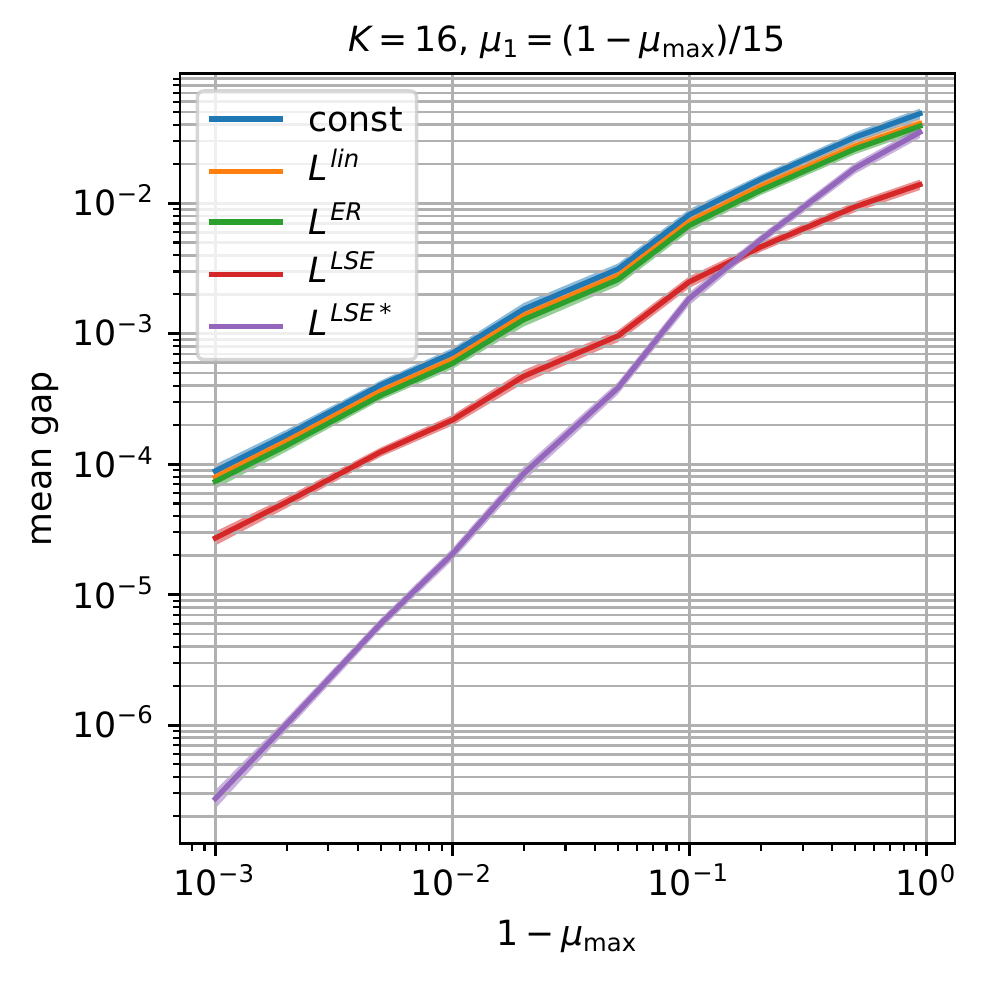}
        \caption{lower bounds, low prob.}
        \label{fig:synth_lbs_d16_low_eps100}
    \end{subfigure}
\caption{Mean gaps of upper bounds (left column) and lower bounds (right column) on softmax output $p_1$ for synthetically generated input regions of width $\epsilon = 1$. Unlike in Figure~\ref{fig:synth}, no ratios of mean gaps are taken. In the top/bottom row, the mean $\mu_1$ of $p_1$ is high/low.}
\label{fig:synth_nonRel}
\end{figure*}

\paragraph{Mean Gaps Without Taking Ratios} Figure~\ref{fig:synth_nonRel} is another version of Figures~\ref{fig:synth_ubs_rel_d16_high_eps100}, \ref{fig:synth_lbs_rel_d16_high_eps100}, \ref{fig:synth_ubs_rel_d16_low_eps100}, \ref{fig:synth_lbs_rel_d16_low_eps100} in which the mean gaps themselves are plotted, without dividing by the mean gap of the constant bound which is now plotted separately. The general pattern is that the mean gaps decrease as $\mu_{\max} \to 1$ ($\mu_1 \to 1$ in the top row of Figure~\ref{fig:synth_nonRel}, $\mu_1 \to 0$ in the bottom row). In Figures~\ref{fig:synth_ubs_d16_high_eps100}, \ref{fig:synth_lbs_d16_high_eps100}, the mean gaps also tend to decrease as $\mu_{\max} \to 0$. The two exceptions are $U^\lin$ in Figure~\ref{fig:synth_ubs_d16_low_eps100}, which is uniformly poor across the $\mu_{\max}$ range, and $L^{\lse}$ in Figure~\ref{fig:synth_lbs_d16_high_eps100}, which is the best lower bound for smaller $\mu_{\max}$ but deterioriates at higher $\mu_{\max}$. These are the same behaviors seen in Figures~\ref{fig:synth_lbs_rel_d16_high_eps100}, \ref{fig:synth_ubs_rel_d16_low_eps100}.

\begin{figure*}[ht]
\centering
    \begin{subfigure}[b]{0.245\textwidth}
        \centering
        \includegraphics[width=\textwidth]{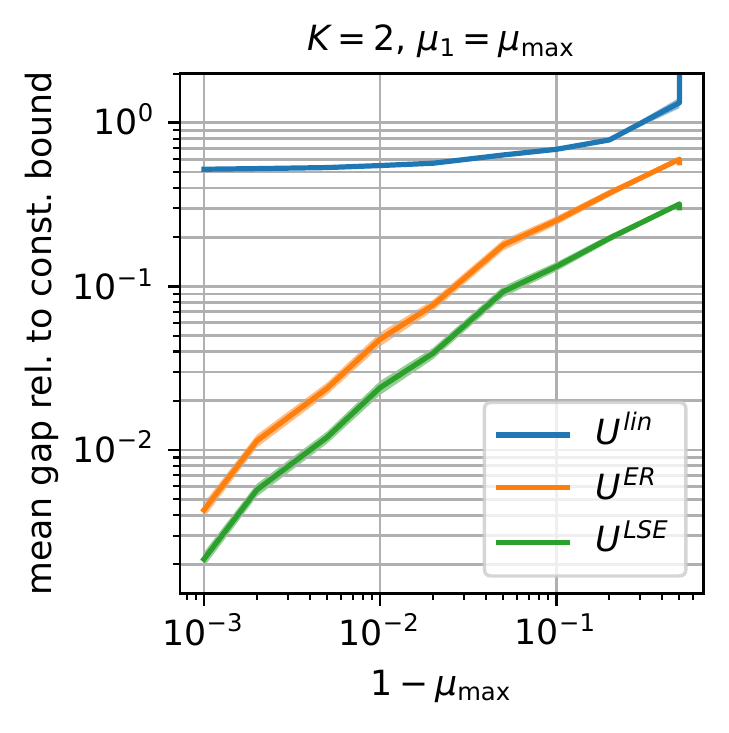}
        \caption{upper bounds, high prob.}
        \label{fig:synth_ubs_rel_d2_high_eps100}
    \end{subfigure}
    \begin{subfigure}[b]{0.245\textwidth}
        \centering
        \includegraphics[width=\textwidth]{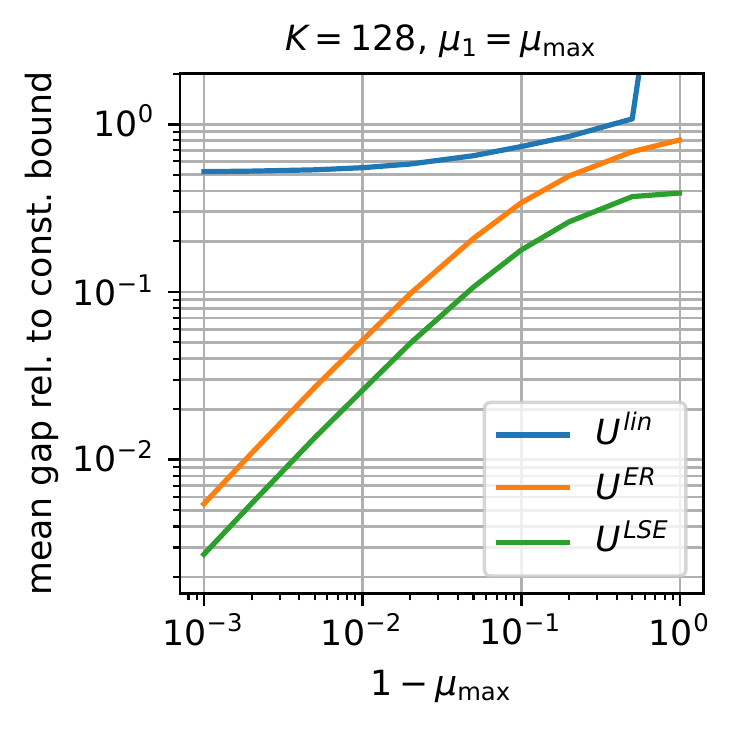}
        \caption{upper bounds, high prob.}
        \label{fig:synth_ubs_rel_d128_high_eps100}
    \end{subfigure}
    \begin{subfigure}[b]{0.245\textwidth}
        \centering
        \includegraphics[width=\textwidth]{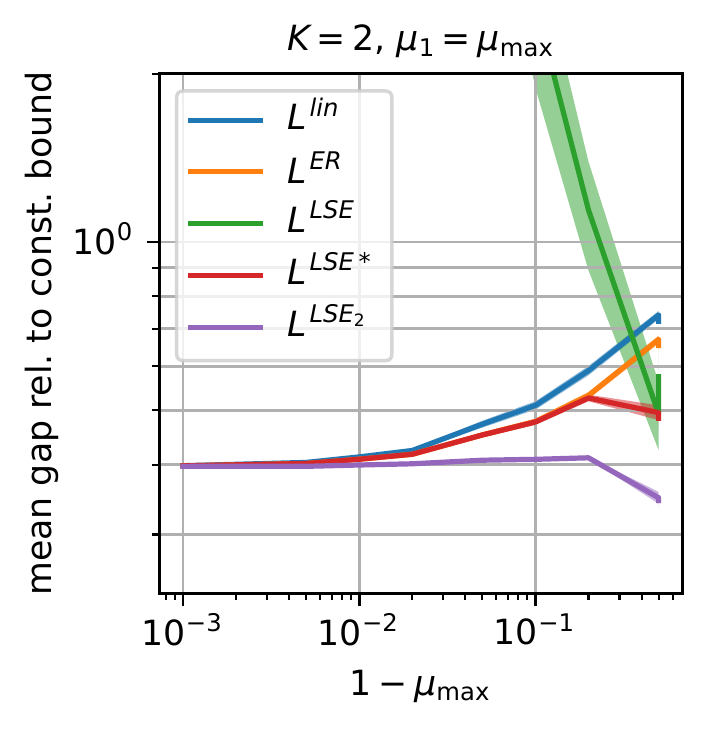}
        \caption{lower bounds, high prob.}
        \label{fig:synth_lbs_rel_d2_high_eps100}
    \end{subfigure}
    \begin{subfigure}[b]{0.245\textwidth}
        \centering
        \includegraphics[width=\textwidth]{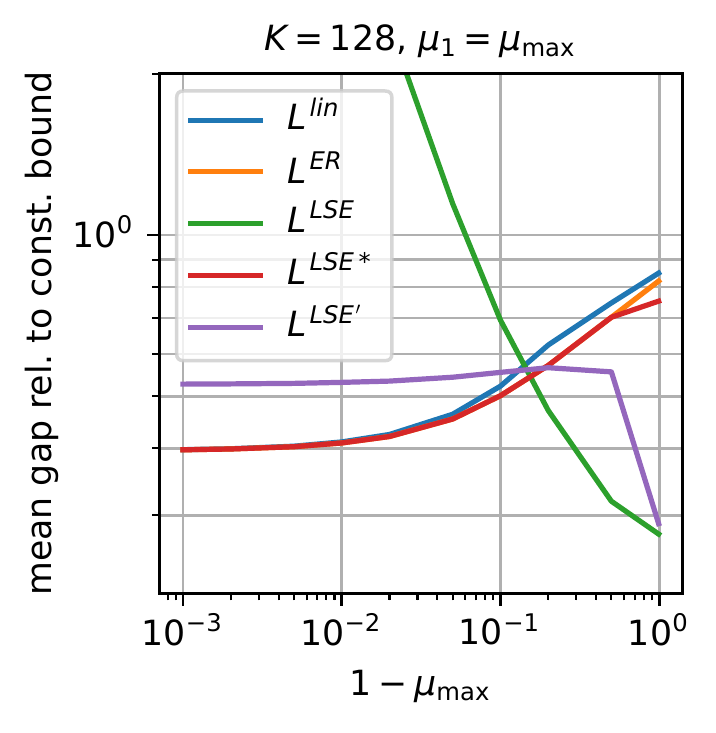}
        \caption{lower bounds, high prob.}
        \label{fig:synth_lbs_rel_d128_high_eps100}
    \end{subfigure}
    \begin{subfigure}[b]{0.245\textwidth}
        \centering
        \includegraphics[width=\textwidth]{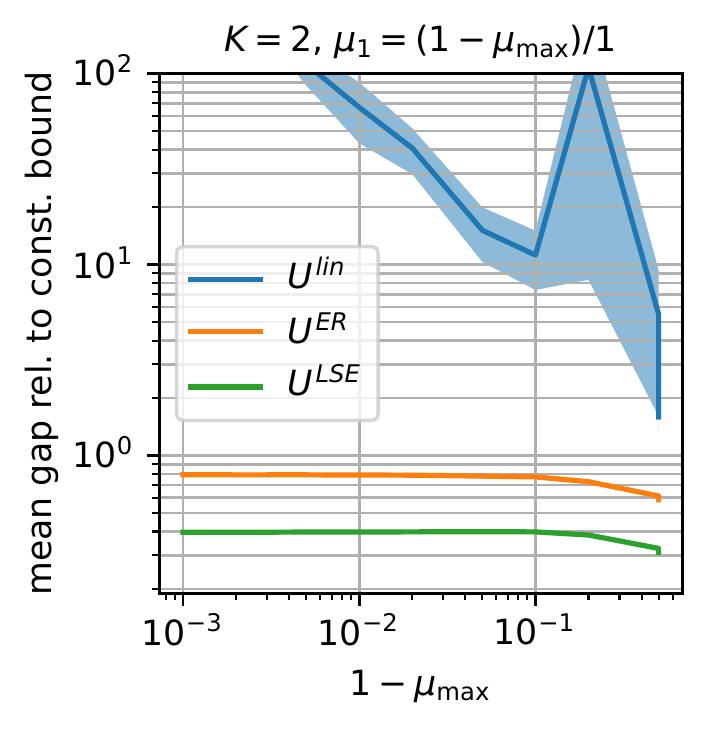}
        \caption{upper bounds, low prob.}
        \label{fig:synth_ubs_rel_d2_low_eps100}
    \end{subfigure}
    \begin{subfigure}[b]{0.245\textwidth}
        \centering
        \includegraphics[width=\textwidth]{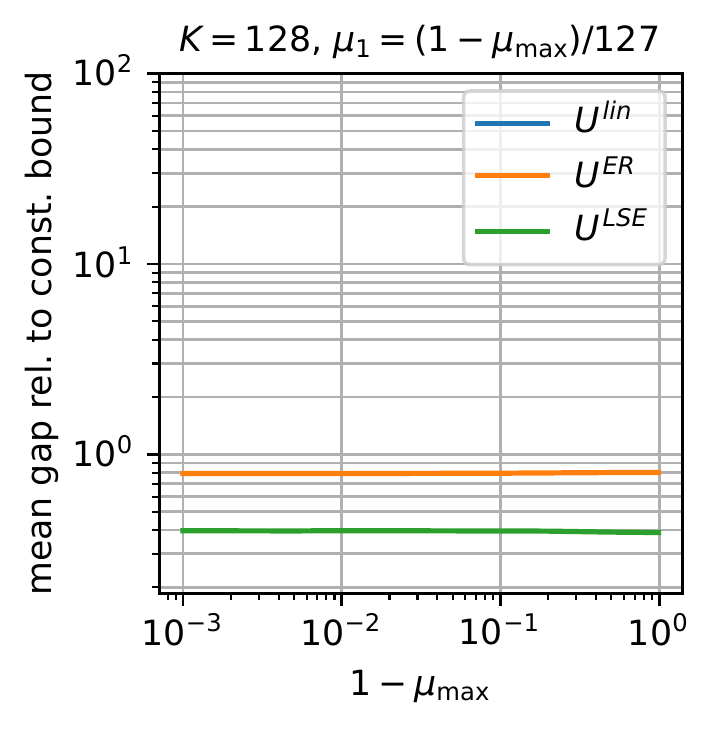}
        \caption{upper bounds, low prob.}
        \label{fig:synth_ubs_rel_d128_low_eps100}
    \end{subfigure}
    \begin{subfigure}[b]{0.245\textwidth}
        \centering
        \includegraphics[width=\textwidth]{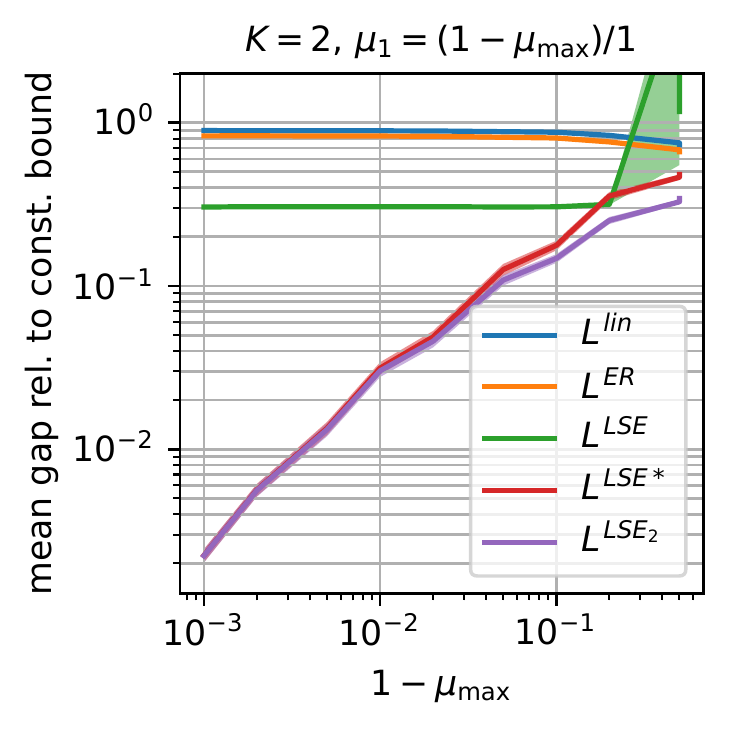}
        \caption{lower bounds, low prob.}
        \label{fig:synth_lbs_rel_d2_low_eps100}
    \end{subfigure}
    \begin{subfigure}[b]{0.245\textwidth}
        \centering
        \includegraphics[width=\textwidth]{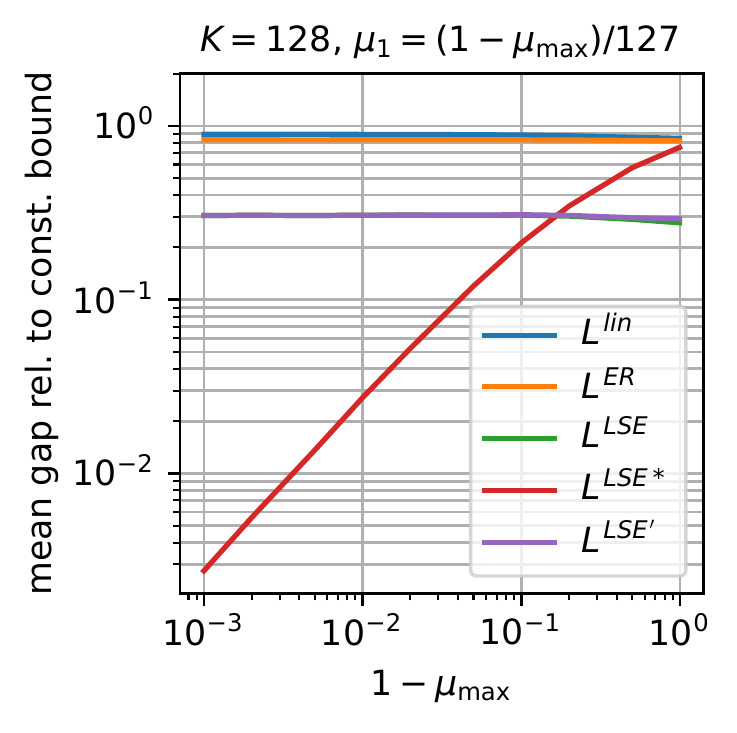}
        \caption{lower bounds, low prob.}
        \label{fig:synth_lbs_rel_d128_low_eps100}
    \end{subfigure}
\caption{Mean gaps of upper bounds (left two columns) and lower bounds (right two columns) on softmax output $p_1$ for synthetically generated input regions of width $\epsilon = 1$ and dimensions $K=2$ and $K=128$. In the top/bottom row, the mean $\mu_1$ of $p_1$ is high/low. For $K = 2$ in Figures~\ref{fig:synth_lbs_rel_d2_high_eps100}, \ref{fig:synth_lbs_rel_d2_low_eps100}, we use \eqref{eqn:LSE_LB2} for the $L^{\lse}$ curve.}
\label{fig:synth_d}
\end{figure*}

\paragraph{Different Values of $K$} Figure~\ref{fig:synth_d} shows versions of Figures~\ref{fig:synth_ubs_rel_d16_high_eps100}, \ref{fig:synth_lbs_rel_d16_high_eps100}, \ref{fig:synth_ubs_rel_d16_low_eps100}, \ref{fig:synth_lbs_rel_d16_low_eps100} with dimensions $K = 2$ and $K = 128$ instead of $K = 16$. In Figure~\ref{fig:synth_ubs_rel_d128_low_eps100}, $U^\lin$ is worse than the constant bound $\olp_1$ by more than a factor of $100$ (the upper limit of the plot is kept at $100$ for consistency). For $K = 2$ in Figures~\ref{fig:synth_lbs_rel_d2_high_eps100}, \ref{fig:synth_lbs_rel_d2_low_eps100}, $L^{\lse_2}$ refers to bound \eqref{eqn:LSE_LB2}, which is provably tighter than $L^\ER$. Figures~\ref{fig:synth_lbs_rel_d2_high_eps100}, \ref{fig:synth_lbs_rel_d2_low_eps100} show that $L^{\lse_2}$ is superior to $L^{\lse}$, $L^{\lse*}$ as well over the entire range of $\mu_{\max}$. For $K = 128$ in Figures~\ref{fig:synth_lbs_rel_d128_high_eps100}, \ref{fig:synth_lbs_rel_d128_low_eps100}, $L^{\lse'}$ is the bound \eqref{eqn:LSE_LBalt} derived in Appendix~\ref{sec:LSE:alt}. As claimed earlier, $L^{\lse'}$ does not improve upon the better of $L^{\lse}$, $L^{\lse*}$ (in Figure~\ref{fig:synth_lbs_rel_d128_low_eps100}, the curve for $L^{\lse}$ largely coincides with that for $L^{\lse'}$ but the former is slightly lower as $\mu_{\max} \to 0$). Aside from the additions of $L^{\lse_2}$ and $L^{\lse'}$, the patterns for the other bounds are similar to those in Figure~\ref{fig:synth}.

\begin{figure*}[t]
\centering
    \begin{subfigure}[b]{0.245\textwidth}
        \centering
        \includegraphics[width=\textwidth]{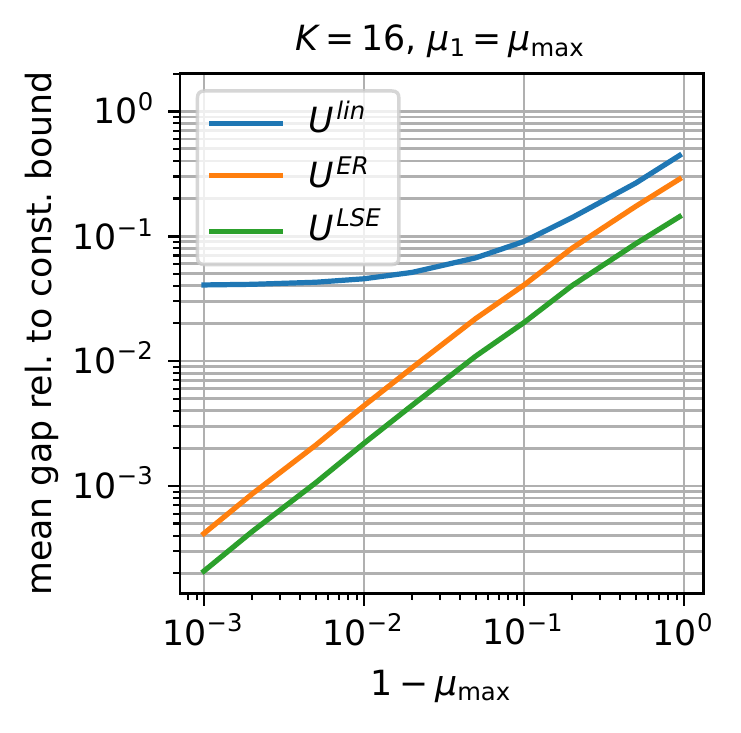}
        \caption{upper bounds, high prob.}
        \label{fig:synth_ubs_rel_d16_high_eps20}
    \end{subfigure}
    \hfill
    \begin{subfigure}[b]{0.245\textwidth}
        \centering
        \includegraphics[width=\textwidth]{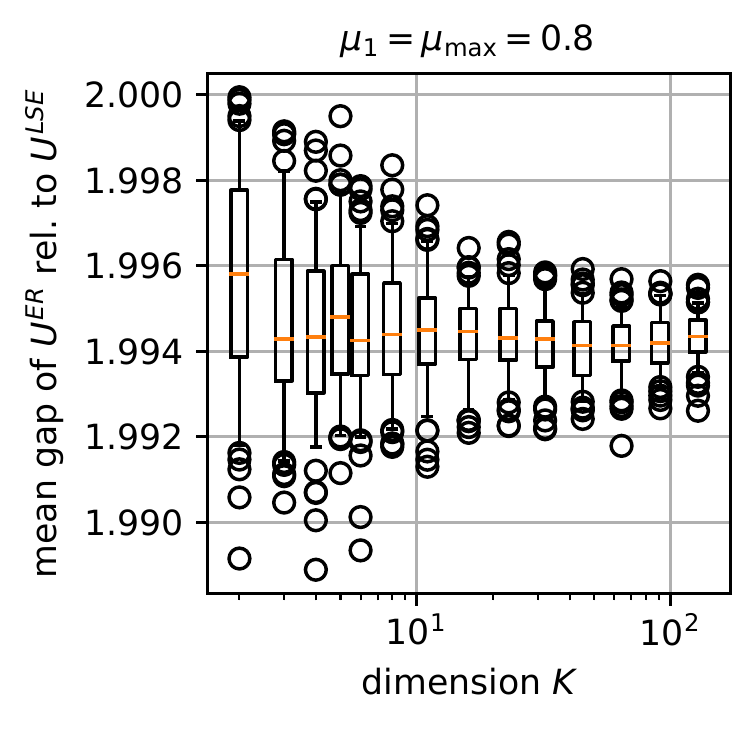}
        \caption{$U^\ER$ vs.~$U^{\lse}$, high prob.}
        \label{fig:synth_ER_u_LSE_u_high0.8_eps20}
    \end{subfigure}
    \hfill
    \begin{subfigure}[b]{0.245\textwidth}
        \centering
        \includegraphics[width=\textwidth]{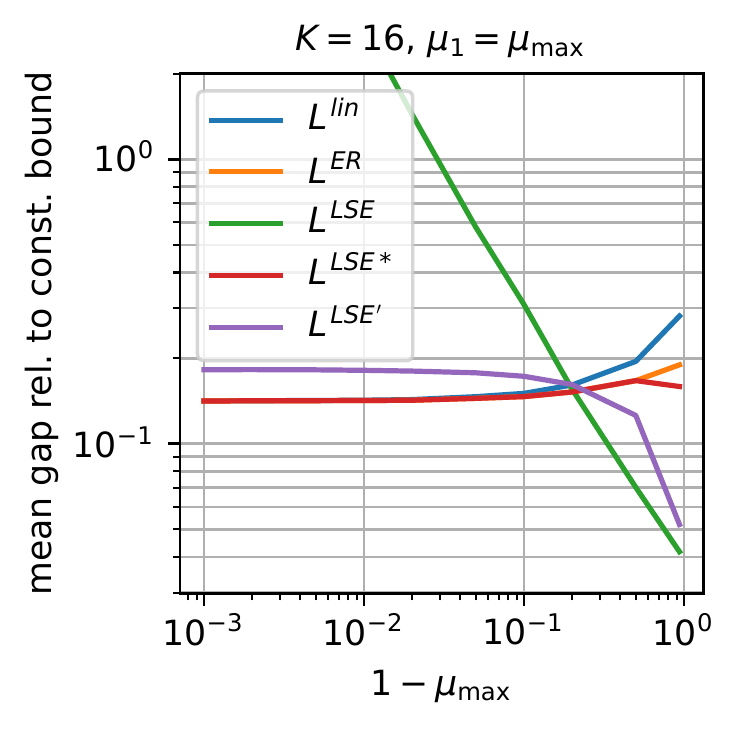}
        \caption{lower bounds, high prob.}
        \label{fig:synth_lbs_rel_d16_high_eps20}
    \end{subfigure}
    \hfill
    \begin{subfigure}[b]{0.245\textwidth}
        \centering
        \includegraphics[width=\textwidth]{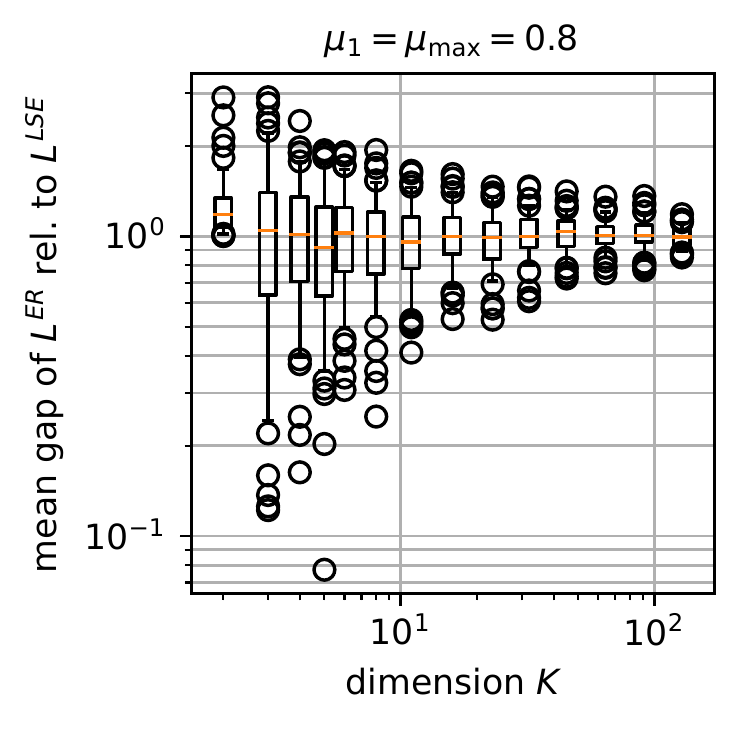}
        \caption{$L^\ER$ vs.~$L^{\lse}$, high prob.}
        \label{fig:synth_ER_l_hybrid_l_high0.8_eps20}
    \end{subfigure}
    \hfill
    \begin{subfigure}[b]{0.245\textwidth}
        \centering
        \includegraphics[width=\textwidth]{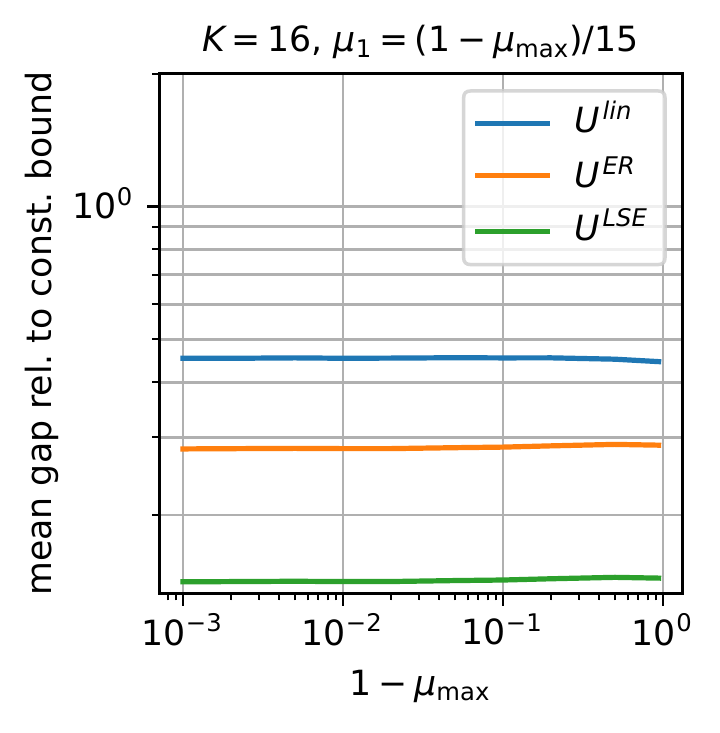}
        \caption{upper bounds, low prob.}
        \label{fig:synth_ubs_rel_d16_low_eps20}
    \end{subfigure}
    \hfill
    \begin{subfigure}[b]{0.245\textwidth}
        \centering
        \includegraphics[width=\textwidth]{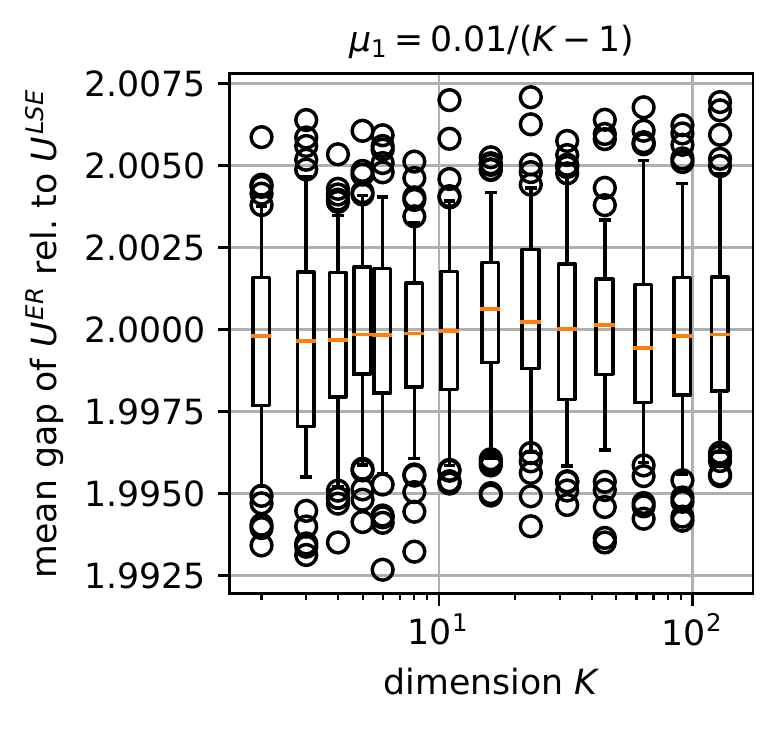}
        \caption{$U^\ER$ vs.~$U^{\lse}$, low prob.}
        \label{fig:synth_ER_u_LSE_u_low0.99_eps20}
    \end{subfigure}
    \hfill
    \begin{subfigure}[b]{0.245\textwidth}
        \centering
        \includegraphics[width=\textwidth]{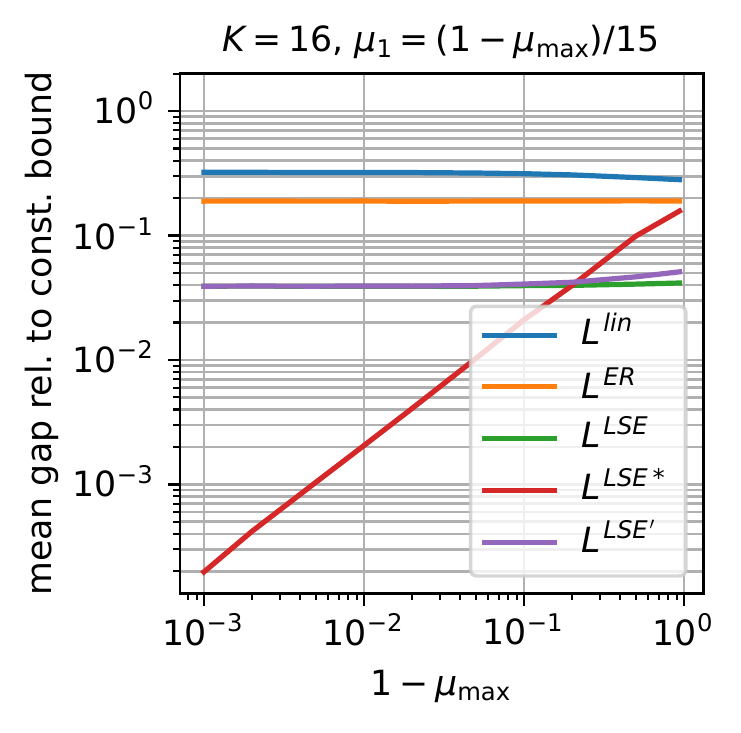}
        \caption{lower bounds, low prob.}
        \label{fig:synth_lbs_rel_d16_low_eps20}
    \end{subfigure}
    \hfill
    \begin{subfigure}[b]{0.245\textwidth}
        \centering
        \includegraphics[width=\textwidth]{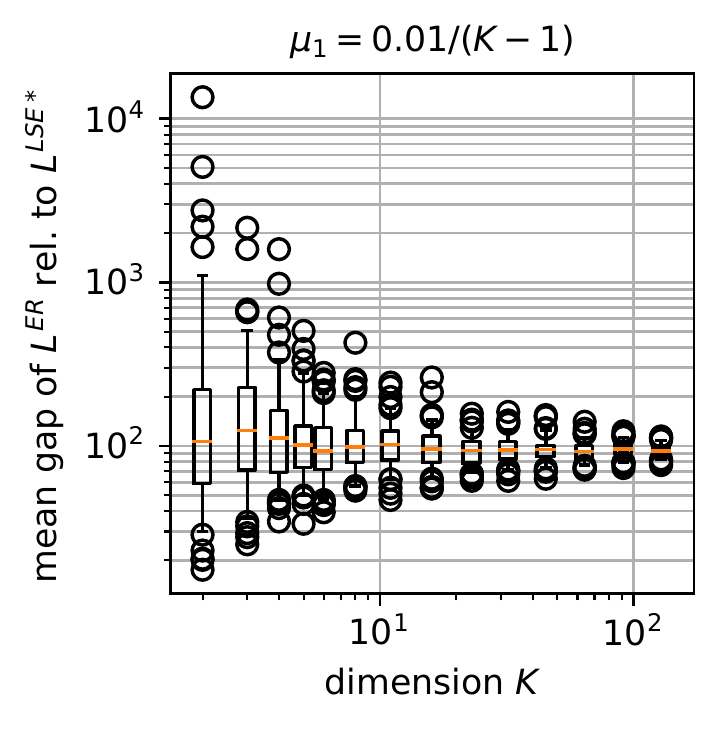}
        \caption{$L^\ER$ vs.~$L^{\lse*}$, low prob.}
        \label{fig:synth_ER_l_hybrid2_l_low0.99_eps20}
    \end{subfigure}
    \hfill
\caption{Mean gaps of upper bounds (left two columns) and lower bounds (right two columns) on softmax output $p_1$ for synthetically generated input regions of width $\epsilon = 0.2$. In the top/bottom row, the mean $\mu_1$ of $p_1$ is high/low.}
\label{fig:synth_eps20}
\end{figure*}

\begin{figure*}[t]
\centering
    \begin{subfigure}[b]{0.245\textwidth}
        \centering
        \includegraphics[width=\textwidth]{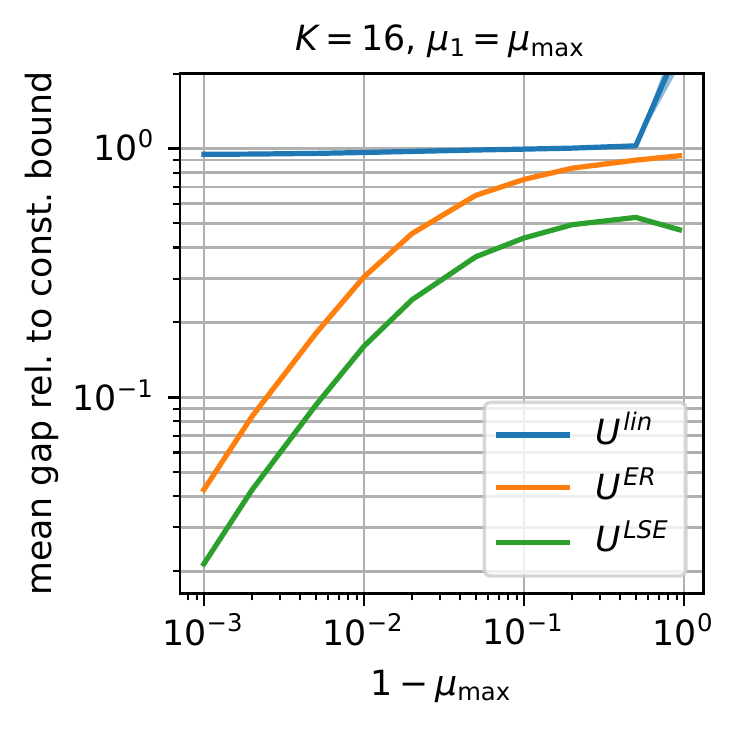}
        \caption{upper bounds, high prob.}
        \label{fig:synth_ubs_rel_d16_high_eps200}
    \end{subfigure}
    \hfill
    \begin{subfigure}[b]{0.245\textwidth}
        \centering
        \includegraphics[width=\textwidth]{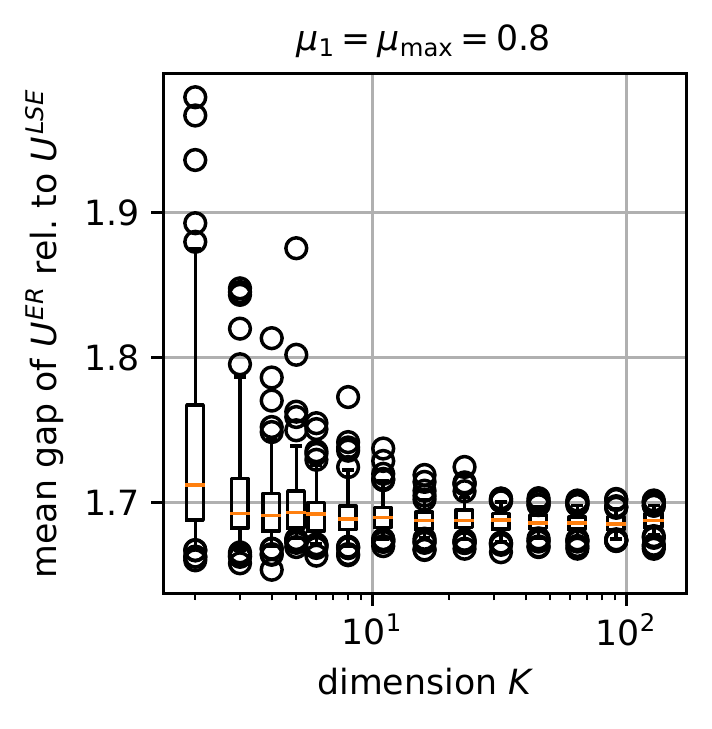}
        \caption{$U^\ER$ vs.~$U^{\lse}$, high prob.}
        \label{fig:synth_ER_u_LSE_u_high0.8_eps200}
    \end{subfigure}
    \hfill
    \begin{subfigure}[b]{0.245\textwidth}
        \centering
        \includegraphics[width=\textwidth]{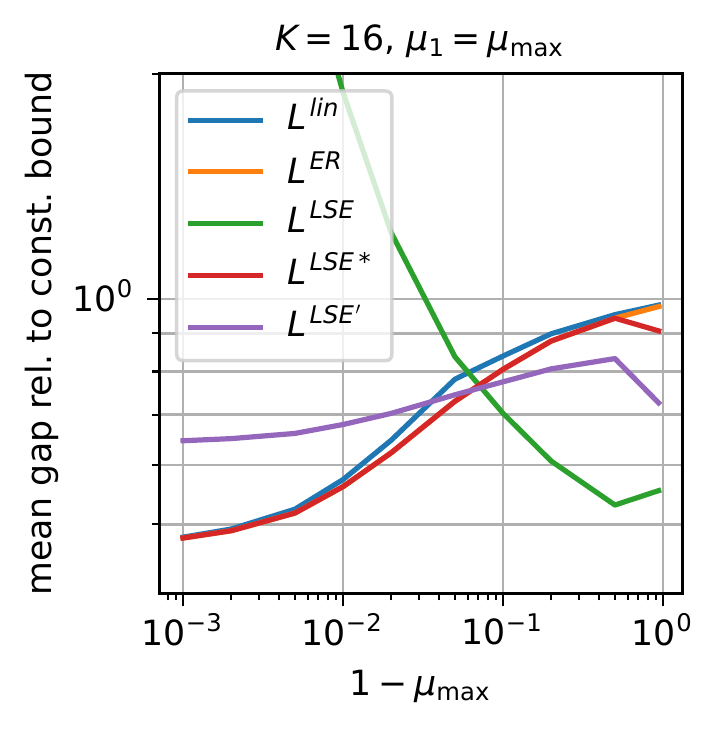}
        \caption{lower bounds, high prob.}
        \label{fig:synth_lbs_rel_d16_high_eps200}
    \end{subfigure}
    \hfill
    \begin{subfigure}[b]{0.245\textwidth}
        \centering
        \includegraphics[width=\textwidth]{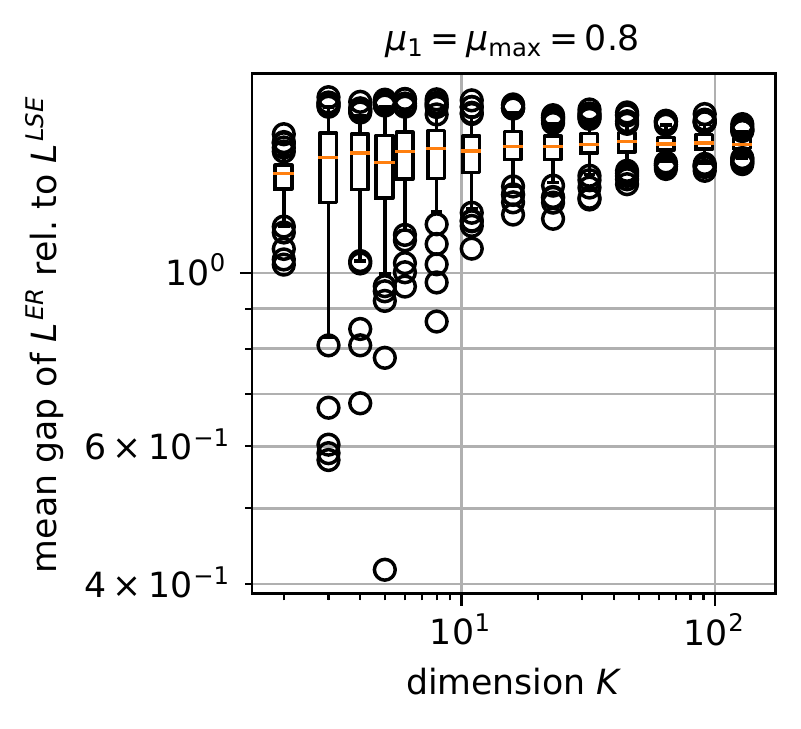}
        \caption{$L^\ER$ vs.~$L^{\lse}$, high prob.}
        \label{fig:synth_ER_l_hybrid_l_high0.8_eps200}
    \end{subfigure}
    \hfill
    \begin{subfigure}[b]{0.245\textwidth}
        \centering
        \includegraphics[width=\textwidth]{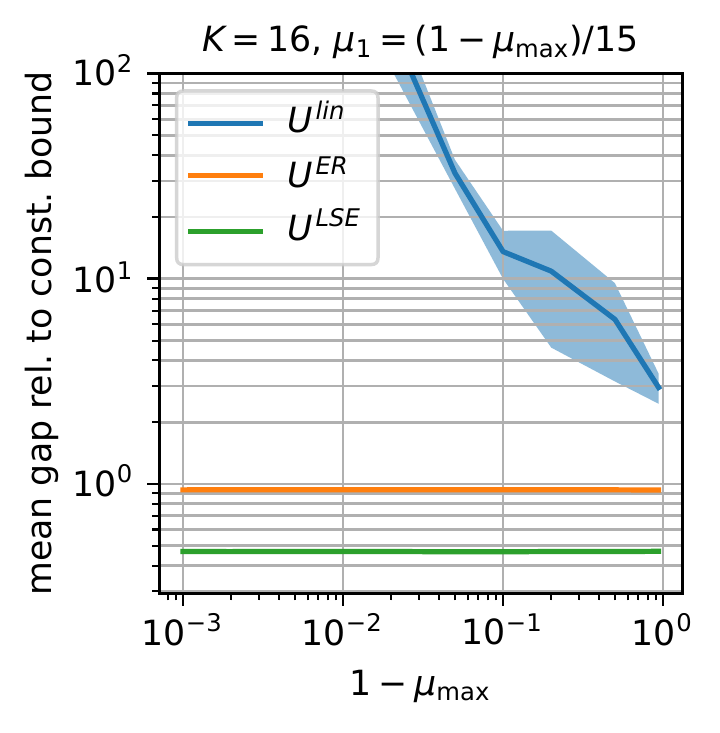}
        \caption{upper bounds, low prob.}
        \label{fig:synth_ubs_rel_d16_low_eps200}
    \end{subfigure}
    \hfill
    \begin{subfigure}[b]{0.245\textwidth}
        \centering
        \includegraphics[width=\textwidth]{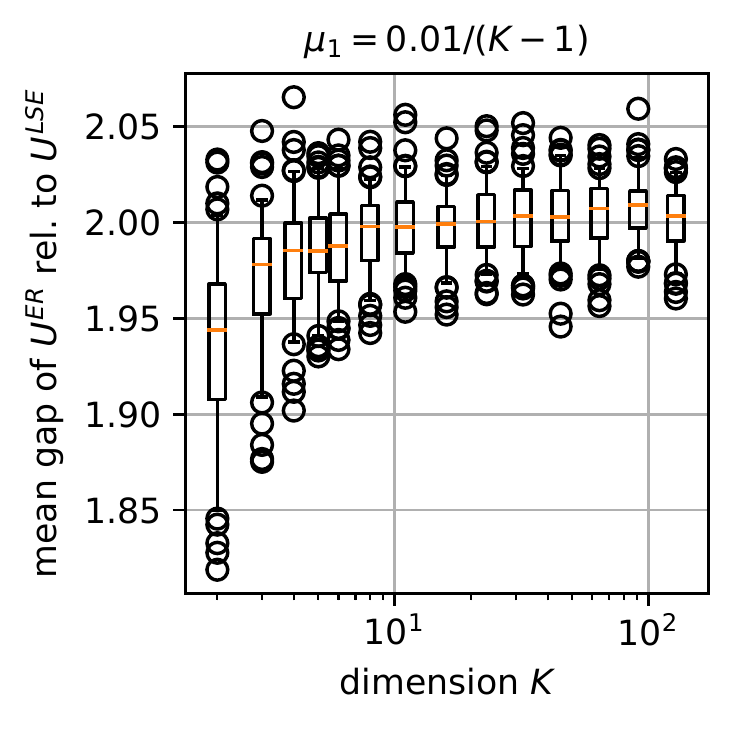}
        \caption{$U^\ER$ vs.~$U^{\lse}$, low prob.}
        \label{fig:synth_ER_u_LSE_u_low0.99_eps200}
    \end{subfigure}
    \hfill
    \begin{subfigure}[b]{0.245\textwidth}
        \centering
        \includegraphics[width=\textwidth]{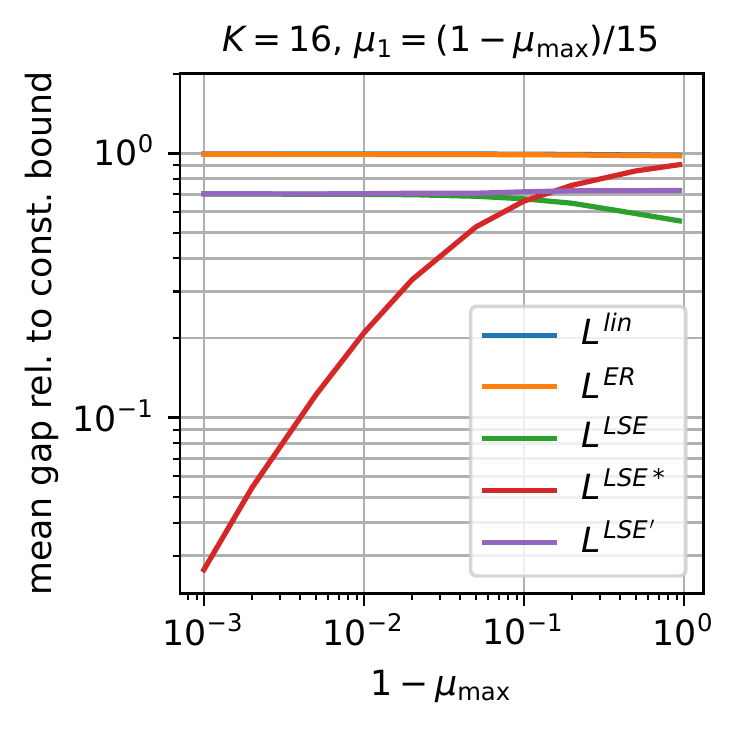}
        \caption{lower bounds, low prob.}
        \label{fig:synth_lbs_rel_d16_low_eps200}
    \end{subfigure}
    \hfill
    \begin{subfigure}[b]{0.245\textwidth}
        \centering
        \includegraphics[width=\textwidth]{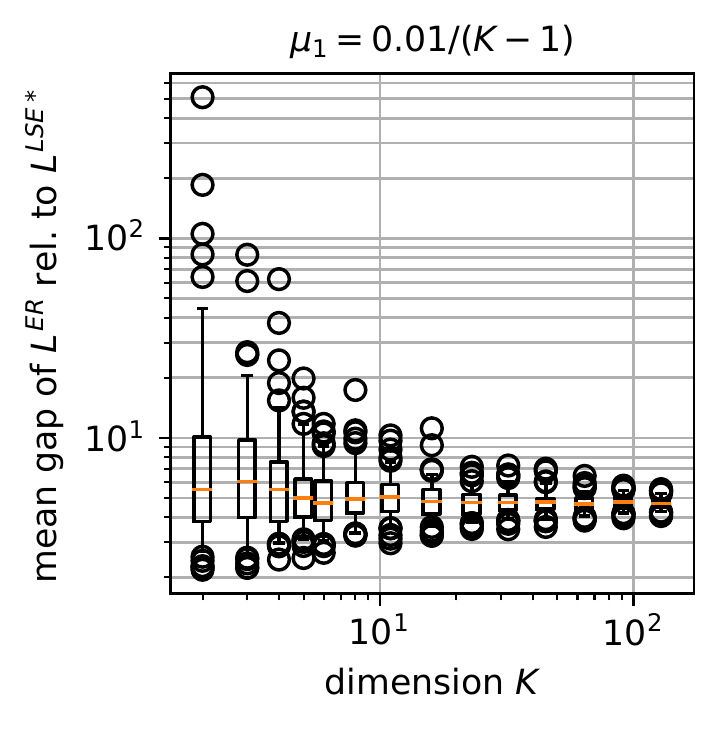}
        \caption{$L^\ER$ vs.~$L^{\lse*}$, low prob.}
        \label{fig:synth_ER_l_hybrid2_l_low0.99_eps200}
    \end{subfigure}
    \hfill
\caption{Mean gaps of upper bounds (left two columns) and lower bounds (right two columns) on softmax output $p_1$ for synthetically generated input regions of width $\epsilon = 2$. In the top/bottom row, the mean $\mu_1$ of $p_1$ is high/low.}
\label{fig:synth_eps200}
\end{figure*}

\paragraph{Different Values of $\epsilon$} Figures~\ref{fig:synth_eps20} and \ref{fig:synth_eps200} are versions of Figure~\ref{fig:synth} with input region width $\epsilon = 0.2$ and $\epsilon = 2$ respectively instead of $\epsilon = 1$. The most notable difference is that for $\epsilon = 0.2$ in Figure~\ref{fig:synth_eps20}, the problem of bounding softmax is easier and the mean gap ratios are generally lower (i.e., improvement over the constant bounds is greater). In particular in Figures~\ref{fig:synth_ubs_rel_d16_high_eps20}, \ref{fig:synth_ubs_rel_d16_low_eps20}, $U^\lin$ is not excessively loose and does improve upon the constant bound $\olp_1$.

\clearpage

\subsection{Looseness of $U^\lin$} We provide an explanation of the looseness of the linear ER bound $U^\lin(x)$ \eqref{eqn:lin_UB} (perhaps not the only explanation). 

Recall that the quantity $t_j$ appearing in \eqref{eqn:lin_UB} is the tangent point \eqref{eqn:tj} chosen to bound the exponential $e^{\tx_j}$ from below by a tangent line. The second term in \eqref{eqn:tj} ensures that the tangent line is non-negative for $\tx_j \in [\tl_j, \tu_j]$. Since $e^{\tx_j}$ is a highly nonlinear function, for large enough intervals $[\tl_j, \tu_j]$ it is likely that we need to set $t_j = \tl_j + 1$ for non-negativity. Suppose then that $t_j = \tl_j + 1$ for all $j = 2,\dots,K$. Then substitution into \eqref{eqn:lin_zLB} yields $\ulq_1^\lin = 1$, and substitution into \eqref{eqn:lin_UB} gives 
\begin{equation}\label{eqn:lin_UB_loose}
    U^\lin(\tx) = 1 - \ulp_1 \sum_{j=2}^K e^{\tl_j+1} (\tx_j - \tl_j).
\end{equation}
When $\tx = \tl$, \eqref{eqn:lin_UB_loose} implies $U^\lin(\tx) = 1$, which is a trivial upper bound. We can also ask when \eqref{eqn:lin_UB_loose} is better (i.e., smaller) than the constant bound $\olp_1$. This occurs when 
\begin{equation}\label{eqn:lin_UB_vs_const_UB}
    \sum_{j=2}^K e^{\tl_j+1} (\tx_j - \tl_j) \geq \frac{1 - \olp_1}{\ulp_1},
\end{equation}
i.e., when $\tx$ is large enough compared to $\tl$ for the above inequality to hold.

The above explanation is consistent with numerical results in Figures~\ref{fig:synth_ubs_rel_d16_high_eps100}, \ref{fig:synth_ubs_rel_d16_low_eps100}, \ref{fig:synth_ubs_rel_d16_high_eps20}, \ref{fig:synth_ubs_rel_d16_low_eps20}, 
\ref{fig:synth_ubs_rel_d16_high_eps200}, \ref{fig:synth_ubs_rel_d16_low_eps200}. When $\epsilon = 0.2$ in Figures~\ref{fig:synth_ubs_rel_d16_high_eps20}, \ref{fig:synth_ubs_rel_d16_low_eps20}, the intervals $[\tl_j, \tu_j]$ are smaller and $U^\lin$ is not as poor as when $\epsilon$ and $[\tl_j, \tu_j]$ are larger and the situation above occurs more frequently. In addition, $U^\lin$ is worse compared to $\olp_1$ in the low $\mu_1$ setting of Figures~\ref{fig:synth_ubs_rel_d16_low_eps100}, \ref{fig:synth_ubs_rel_d16_low_eps20},
\ref{fig:synth_ubs_rel_d16_low_eps200} than in the high $\mu_1$ setting. In this case, both $\ulp_1$ and $\olp_1$ tend to be small, the right-hand side of \eqref{eqn:lin_UB_vs_const_UB} is large, and \eqref{eqn:lin_UB_vs_const_UB} may not hold for any $\tx \in [\tl, \tu]$.

\section{FULL FORMULATION OF UNCERTAINTY ESTIMATION SCORE MAXIMIZATION}
\label{sec:appl:UQ:formulation}

We consider an ensemble of $M$ feedforward NNs, each consisting of $L$ hidden layers with ReLU activations. Let $x^{\ell,m}$ denote the neurons in layer $\ell$ of network $m$, $\ell=0,\dots,L$, $m=1,\dots,M$, before affine transformation is applied. The input to the ensemble corresponds to $\ell=0$ and is the same for all networks: $x^{0,m} = x^0$. Let $W^{\ell,m}$ and $b^{\ell,m}$ be the weights and biases of the affine transformation in layer $\ell$ of network $m$, and $z^{\ell,m}$ be the output of the affine transformation. $z^{L,m}$ is the set of logits from network $m$ and is the input to a softmax function with probabilities $p^{m}$ as output. The output probabilities of the ensemble are the averages of the network probabilities, $p = (1/M) \sum_m p^{m}$. We assume that we have lower and upper bounds $l^{\ell,m}$, $u^{\ell,m}$ on each $z^{\ell,m}$. In our experiment, these are obtained using the CROWN/DeepPoly~\citep{zhang2018efficient,singh2019abstract} abstract interpretation.

Below we give the full set of constraints in the score maximization problem \eqref{eqn:advScore} for verification of uncertainty estimation.

\paragraph{Input} We consider $\ell_\infty$ perturbations up to radius $\epsilon$ of the given input $x^*$ (i.e., $\mathcal{B}_\infty(x^*, \epsilon)$ in \eqref{eqn:advScore}). This can be expressed as the following linear constraints on $x^0$:
\begin{equation}\label{eqn:inputBounds}
    x^* - \epsilon \leq x^0 \leq x^* + \epsilon.
\end{equation}

\paragraph{Hidden Layers} Given bounds $l^{\ell,m}$, $u^{\ell,m}$ on the pre-activation neurons $z^{\ell,m}$ for layers $l = 0,\dots,L-1$, we can partition the post-activation neurons $x^{\ell+1,m}$ into three sets $\cI, \cA, \cU$ (we may regard all components of $x^0$ as belonging to $\cA$):
\begin{enumerate}
    \item \textit{Inactive}, $\cI = \{j : u^{\ell,m}_j \leq 0\}$: In this case, $x^{\ell+1,m}_j = 0$ and can be dropped as an input to the next affine transformation.
    \item \textit{Active}, $\cA = \{j : l^{\ell,m}_j \geq 0\}$: This implies that $x^{\ell+1,m}_j = z^{\ell,m}_j$ and $x^{\ell+1,m}_j$ is an affine function of $x^{\ell,m}$. With $x^{\ell+1,m}_\cA$ denoting the subvector of $x^{\ell+1,m}$ indexed by $\cA$ and using similar notation for other vectors and matrices, these affine functions can be written as
    \begin{equation}
        x^{\ell+1,m}_\cA = W^{\ell,m}_{\cA,\cA\cup\cU} x^{\ell,m}_{\cA\cup\cU} + b^{\ell,m}_\cA, \qquad \ell = 0,\dots,L-1, \; m = 1,\dots,M.
    \end{equation}
    \item \textit{Unstable}, $\cU = \{j : l^{\ell,m}_j < 0 < u^{\ell,m}_j\}$: Here the ReLU function remains nonlinear and we approximate it using the triangular linear relaxation of \citet{ehlers2017formal}. This gives us the following constraints:
    \begin{align}
        z^{\ell,m}_\cU &= W^{\ell,m}_{\cU,\cA\cup\cU} x^{\ell,m}_{\cA\cup\cU} + b^{\ell,m}_\cU,\\
        x^{\ell+1,m}_\cU &\geq z^{\ell,m}_\cU, \qquad x^{\ell+1,m}_\cU \geq 0,\\
        x^{\ell+1,m}_j &\leq \frac{u^{\ell,m}_j}{u^{\ell,m}_j - l^{\ell,m}_j} \left( z^{\ell,m}_j - l^{\ell,m}_j \right), \quad j \in \cU, \; \ell = 0,\dots,L-1, \; m = 1,\dots,M.
    \end{align}
\end{enumerate}

\paragraph{Logits} These are given by affine transformation of the last hidden layer:
\begin{equation}
    z^{L,m} = W^{L,m}_{\cdot,\cA\cup\cU} x^{L,m}_{\cA\cup\cU} + b^{L,m}.
\end{equation}

\paragraph{Probabilities} We impose the simplex constraint in \eqref{eqn:simplex} on the probabilities from each network:
\begin{equation}\label{eqn:simplex_UQ}
    \sum_{k=1}^K p^{m}_k = 1, \qquad p^{m}_k \geq 0, \qquad k=1,\dots,K, \; m = 1,\dots,M.
\end{equation}

\paragraph{Objective Function} Equation~\eqref{eqn:NLL} is used for the NLL scoring rule and \eqref{eqn:BrierUB} for Brier score, with $p = (1/M) \sum_{m=1}^M p^{m}$ in both cases. The fact that \eqref{eqn:BrierUB} is an upper bound on the Brier score can be seen from the chordal upper bound on the function $p_k^2$ over the interval $[\ulp_k, \olp_k]$. This chordal upper bound can be simplified to
\[
    p_k^2 \leq (\ulp_{k} + \olp_{k}) p_k - \ulp_k \olp_k.
\]

\paragraph{Softmax} We use the bounds developed in this work to relate logits to probabilities:
\begin{subequations}
\begin{align}
    p^m_k &\geq L_k\left(z^{L,m}\right),\label{eqn:L_UQ}\\
    p^m_k &\leq U_k\left(z^{L,m}\right),\label{eqn:U_UQ}
\end{align}
\end{subequations}
where the subscript $k$ in $L_k$, $U_k$ refers to the bound for the $k$th softmax output (all the bounds presented in the paper were for $k=1$). In our experiments, $L_k \in \{L^\lin, L^\ER, L^{\lse*}\}$ and $U_k \in \{U^\lin, U^{\lse}\}$ as discussed in Section~\ref{sec:appl:UQ}.  However, because of the form of the objective functions \eqref{eqn:NLL}, \eqref{eqn:BrierUB}, only one of \eqref{eqn:L_UQ}, \eqref{eqn:U_UQ} is needed for each $k$. Specifically, for $k = y^*$, the probability $p_{y^*}$ is minimized in both objective functions\footnote{In the case of Brier score \eqref{eqn:BrierUB}, the coefficient of $p_{y^*}$ is $(-2 + \ulp_k + \olp_k) \leq 0$ and hence $p_{y^*}$ is minimized as in \eqref{eqn:NLL}.}, whereas for $k \neq y^*$, $p_k$ is maximized\footnote{In the case of NLL, the maximization is implicit because $p_{y^*}$ is minimized and \eqref{eqn:simplex_UQ} couples $p_{y^*}$ to the other $p_k$'s.}. Therefore only \eqref{eqn:L_UQ} is used for $k = y^*$ and \eqref{eqn:U_UQ} for $k \neq y^*$.

In addition, we also impose the constant bounds 
\begin{subequations}\label{eqn:const_UQ}
\begin{align}
    p^m_k &\geq \ulp^m_k,\label{eqn:const_LB_UQ}\\
    p^m_k &\leq \olp^m_k,\label{eqn:const_UB_UQ}
\end{align}
\end{subequations}
where again only \eqref{eqn:const_LB_UQ} is used for $k = y^*$ and \eqref{eqn:const_UB_UQ} for $k \neq y^*$. In the case of the linear pair $(L^\lin, U^\lin)$, constraints \eqref{eqn:const_UQ} help considerably because $U^\lin$ in particular is sometimes not better than $\olp^m_k$ (recall for example Figure~\ref{fig:synth_ubs_rel_d16_low_eps100}). For the nonlinear pairs $(L^\ER, U^{\lse})$, $(L^{\lse*}, U^{\lse})$, \eqref{eqn:const_UQ} can be helpful in improving the numerical precision of the optimal objective value.

\paragraph{Summary} In summary, problem \eqref{eqn:advScore} is subject to constraints \eqref{eqn:inputBounds}--\eqref{eqn:const_UQ}, where \eqref{eqn:L_UQ}, \eqref{eqn:const_LB_UQ} are used for $k = y^*$ and \eqref{eqn:U_UQ}, \eqref{eqn:const_UB_UQ} for $k \neq y^*$.

\section{DETAILS ON MODEL ARCHITECTURES AND TRAINING \label{sec:arch}}

We describe the details of the deep ensembles deployed in predictive uncertainty estimation of Section~\ref{sec:appl:UQ} as well as the self-attention models used in Section~\ref{sec:appl:attention}.

\subsection{Deep Ensembles}

We follow the layer structures of the deep ensemble models for the MNIST dataset as in Section~3.4 of \cite{lakshminarayanan2017simple}. While the authors did not opensource the accompanying implementation codes for this paper, we are advised that the \textsf{uncertainty-baselines}\footnote{Access via GitHub repository \url{https://github.com/google/uncertainty-baselines}} Python package can be utilised to construct the training pipeline. In particular, we look into the \textsf{./baselines/mnist/} directory and train the following MNIST deep ensemble models in Table~\ref{tab:mnist-arch}. Specifically, we have two ensembles, \textsf{MNIST} and \textsf{MNIST-large}, each containing 5 networks with identical layer structure yet different weights and biases after the training process. Overall, networks of the \textsf{MNIST} ensemble reach $\sim 94.5\%$ test accuracy, while the \textsf{MNIST-large} ensemble networks achieve $\sim 97.95\%$. Table~\ref{tab:UQ} in Section~\ref{sec:appl:UQ} shows verification results for \textsf{MNIST}, while results for \textsf{MNIST-large} are in Table~\ref{tab:UQ_large} in Appendix~\ref{sec:appl:UQ:large}.  We mention that, as recommended by \cite{lakshminarayanan2017simple}, the \textsf{MNIST-large} models are adversarially trained using the Projected Gradient Descent~\citep{pgd} attacking method from the Adversarial Robustness Toolbox\footnote{Access via GitHub repository \url{https://github.com/Trusted-AI/adversarial-robustness-toolbox}} Python library.
Similarly, we adversarially train an ensemble comprising 5 networks on the \textsf{CIFAR-10} dataset, and report their verification results in Table~\ref{tab:UQ_cifar} of Section~\ref{sec:appl:UQ}. The structure of these \textsf{CIFAR-10} models is in Table~\ref{tab:cifar10-arch}, and their test accuracy varies from $28.49\%$ to $42.71\%$.

\begin{table}[h]
    \caption{Architecture for the \textsf{MNIST} (left) and \textsf{MNIST-large} (right) deep ensemble models.} 
    \label{tab:mnist-arch}
    \centering		
    \renewcommand{\arraystretch}{1.3}
    \begin{tabular}{c|c|c}
        \hline \hline
        Layer Type & Parameter & Activation \\
        \hline
        Input & $28 \times 28 \times 1$ & -- \\
        Flatten & $784$ & -- \\
        Fully Connected & $784 \times 10$ & ReLU \\
        Fully Connected & $10 \times 10$ & ReLU \\
        Fully Connected & $10 \times 10$ & Softmax \\
        \hline \hline
    \end{tabular}
    \qquad\qquad
    \begin{tabular}{c|c|c}
        \hline \hline
        Layer Type & Parameter & Activation \\
        \hline
        Input & $28 \times 28 \times 1$ & -- \\
        Flatten & $784$ & -- \\
        Fully Connected & $784 \times 100$ & ReLU \\
        Fully Connected & $100 \times 100$ & ReLU \\
        Fully Connected & $100 \times 100$ & ReLU \\
        Fully Connected & $100 \times 10$ & Softmax \\
        \hline \hline
    \end{tabular}
\end{table}

\begin{table}[h]
    \caption{Architecture for the \textsf{CIFAR-10} deep ensemble model.} 
    \label{tab:cifar10-arch}
    \centering		
    \renewcommand{\arraystretch}{1.3}
    \begin{tabular}{c|c|c}
        \hline \hline
        Layer Type & Parameter & Activation \\
        \hline
        Input & $32 \times 32 \times 3$ & -- \\
        Flatten & $3072$ & -- \\
        Fully Connected & $3072 \times 20$ & ReLU \\
        Fully Connected & $20 \times 20$ & ReLU \\
        Fully Connected & $20 \times 20$ & ReLU \\
        Fully Connected & $20 \times 10$ & Softmax \\
        \hline \hline
    \end{tabular}
\end{table}

\subsection{Self-Attention Models}

We adapt the self-attention mechanism proposed in \cite{vaswani2017attention}, where essentially encoders and decoders comprising attention blocks are used in processing sentences, into classifying the MNIST and SST-2 datasets. The layer structure of our MNIST self-attention model is outlined in Table~\ref{tab:attention-arch}. Specifically, we train $3$ models with increasing dimensions of the self-attention block in terms of the number of attention heads and the size of each head: \attSmall, \attMed, \attBig with test accuracy $90.25\%$, $97.41\%$, and $98.28\%$, respectively. Similarly, we deploy adversarial training by using the Projected Gradient Descent~\citep{pgd} attacking method from the Adversarial Robustness Toolbox Python library. As for the SST-2 sentiment analysis dataset, we adversarially train the self-attention model in Table~\ref{tab:attention-arch-sst}. While the self-attention block is the same as \attSmall for MNIST, it has the embedding layer to accommodate text inputs and the sigmoid function of the last layer to produce sentiment, i.e., either \textsf{positive} or \textsf{negative}. The test accuracy of this model is $75.34\%$.

\begin{table}[h]
    \caption{Architecture for the MNIST self-attention models: \attSmall ($90.25\%$), \attMed ($97.41\%$), and \attBig ($98.28\%$).} 
    \label{tab:attention-arch}
    \centering		
    \renewcommand{\arraystretch}{1.3}
    \begin{tabular}{c|c|c}
        \hline \hline
        Layer Type & Parameter Size & Activation \\
        \hline
        Input & $28 \times 28 \times 1$ & -- \\
        Flatten &  -- & -- \\
        Fully Connected & \attSmall: $16$; \attMed: $144$; \attBig: $256$ & ReLU \\
        Reshape & \attSmall: $(2, 8)$; \attMed: $(4, 36)$; \attBig: $(4, 64)$ & -- \\
        \multirow{3}{*}{MultiHeadAttention} & \attSmall: num\_heads=$2$, key\_dim=$4$ \\
        & \attMed: num\_heads=$3$, key\_dim=$12$ & -- \\
        & \attBig: num\_heads=$4$, key\_dim=$16$ \\
        Flatten &  -- & -- \\
        Fully Connected & $10$ & Softmax \\
        \hline \hline
    \end{tabular}
\end{table}

\begin{table}[h]
    \caption{Architecture for the SST-2 self-attention model ($75.34\%$).} 
    \label{tab:attention-arch-sst}
    \centering		
    \renewcommand{\arraystretch}{1.3}
    \begin{tabular}{c|c|c}
        \hline \hline
        Layer Type & Parameter Size & Activation \\
        \hline
        Input & $50$ & -- \\
        Embedding & vocab\_size=$4000$, embedding\_dim=$16$, input\_length=$50$ & -- \\
        Flatten &  -- & -- \\
        Fully Connected & $16$ & ReLU \\
        Reshape & $(2, 8)$ & -- \\
        MultiHeadAttention & num\_heads=$2$, key\_dim=$4$ & -- \\
        Flatten &  -- & -- \\
        Fully Connected & $1$ & Sigmoid \\
        \hline \hline
    \end{tabular}
\end{table}

\section{ADDITIONAL UNCERTAINTY ESTIMATION RESULTS}

\subsection{Lower Bound on UQ Scores Computed by PGD Attack}
\label{sec:appl:UQ:PGD}

We perform projected gradient descent (PGD) attack to obtain a lower bound of the UQ scores within the given perturbation bound. The results are shown in Table~\ref{tab:UQ_with_attack}.

\begin{table}[h]
\setlength\tabcolsep{3 pt}
\caption{Upper Bounds on Uncertainty Estimation Scores Using Different Softmax Bounds} \label{tab:UQ_with_attack}
\begin{center}
\begin{tabular}{llllll}
\textbf{Score (Clean)}   &$\epsilon$ & PGD & $L^\lin, U^\lin$ & $L^\ER, U^{\lse}$ & $L^{\lse*}, U^{\lse}$ \\
\hline 
NLL (0.105)   & 2/256 & 0.202 & 0.265 & 0.261 & \textbf{0.251} \\
        & 3/256 & 0.285 & 0.442 & 0.433 & \textbf{0.420} \\
        & 4/256 & 0.407 & 0.726 & 0.697 & \textbf{0.690} \\ \hline
Brier (0.048)  & 2/256 & 0.061 & 0.138 & 0.134 & \textbf{0.131} \\
        & 3/256 & 0.073 & 0.244 & 0.235 & \textbf{0.234} \\
        & 4/256 & 0.104 & 0.417 & \textbf{0.403} & \textbf{0.403} \\ \hline
\end{tabular}
\end{center}
\end{table}

\subsection{Verification of a Larger Deep Ensemble}
\label{sec:appl:UQ:large}

Table~\ref{tab:UQ_large} shows upper bounds on expected uncertainty estimation scores for the \textsf{MNIST-large} ensemble in the same manner as Table~\ref{tab:UQ} for \textsf{MNIST} in Section~\ref{sec:appl:UQ}. The table also includes results for two larger perturbation radii, 5/256 and 6/256 ($0.020$ and $0.024$ in $[0, 1]$-normalized units). In the case of \textsf{MNIST-large}, the lower bound $L^{\lse*}$ appears to coincide with $L^\ER$ for all $100$ test instances, and thus the two rightmost columns in Table~\ref{tab:UQ_large} are identical. Nevertheless, there is still consistent improvement in going from linear to nonlinear bounds.

\begin{table}[ht]
\caption{Upper Bounds on Uncertainty Estimation Scores for a Larger Deep Ensemble Using Different Softmax Bounds} \label{tab:UQ_large}
\begin{center}
\begin{tabular}{llllll}
\textbf{Score (Clean)}   &$\epsilon$ & PGD & $L^\lin, U^\lin$ & $L^\ER, U^{\lse}$ & $L^{\lse*}, U^{\lse}$ \\
\hline 
NLL (0.0150)   & 2/256  & 0.0196 & 0.0212 & 0.0210 & 0.0210 \\
        & 3/256  & 0.0222 & 0.0263 & 0.0257 & 0.0257 \\
        & 4/256  & 0.0252 & 0.0339 & 0.0326 & 0.0326 \\ 
        & 5/256  & 0.0286 & 0.0449 & 0.0430 & 0.0430 \\ 
        & 6/256  & 0.0325 & 0.0611 & 0.0585 & 0.0585 \\ 
        \hline
Brier (0.00373)  & 2/256 & 0.00584 & 0.00677 & 0.00659 & 0.00659 \\
        & 3/256 & 0.00707 & 0.00961 & 0.00921 & 0.00921 \\
        & 4/256 & 0.00847 & 0.01418 & 0.01350 & 0.01350 \\ 
        & 5/256 & 0.00976 & 0.02161 & 0.02065 & 0.02065 \\
        & 6/256 & 0.01125 & 0.03336 & 0.03206 & 0.03206 \\ 
        \hline
\end{tabular}
\end{center}
\end{table}

\subsection{UQ Scores Computed by Considering Each Network Separately}
\label{sec:appl:UQ:separate}

Since the output of the ensemble models we study are average of the outputs of the individual models,  
another way to compute the UQ scores for an ensemble models is to computing the scores for each individual model and take the average. Note that this is not equivalent to analyzing the ensemble model as a whole, because the constraint that each model always take the same input is relaxed and the resulting average score is an over-approximation. One benefit though is that analyzing each individual model is computationally more efficient than analyzing the ensemble model as a whole.

\begin{table}[h]
\setlength\tabcolsep{3 pt}
\caption{Upper Bounds on Uncertainty Estimation Scores Using Different Softmax Bounds} \label{tab:UQ_separate}
\begin{center}
\begin{tabular}{lllll}
\textbf{Score (Clean)}   &$\epsilon$ & $L^\lin, U^\lin$ & $L^\ER, U^{\lse}$ & $L^{\lse*}, U^{\lse}$ \\
\hline 
NLL (0.105)   & 2/256 & 0.270 & \textbf{0.263} & \textbf{0.263} \\
        & 3/256 & 0.448 & \textbf{0.435} & 0.436 \\
        & 4/256 & 0.732 & \textbf{0.712} & \textbf{0.712} \\ \hline
Brier (0.048)  & 2/256 & 0.139 & \textbf{0.136} & \textbf{0.136} \\
        & 3/256 & 0.245 & \textbf{0.240} & \textbf{0.240} \\
        & 4/256 & 0.418 & \textbf{0.412} & \textbf{0.412} \\ \hline
\end{tabular}
\end{center}
\end{table}

\begin{table}[h]
\setlength\tabcolsep{2 pt}
\caption{Average runtime in seconds} 
\begin{center}
\begin{tabular}{llll}
\textbf{Dataset} & $L^\lin, U^\lin$ & $L^\ER, U^{\lse}$ & $L^{\lse*}, U^{\lse}$ \\
\hline 
Combined   & 10.9 & 91.6 & 92.4 \\
Separate & 3.36 & 43.22 & 42.49 \\
\hline
\end{tabular}
\end{center}
\end{table}

